\pdfoutput=1
\documentclass[11pt]{article}

\usepackage[final]{acl}

\usepackage{times}
\usepackage{latexsym}

\usepackage[T1]{fontenc}
\usepackage[utf8]{inputenc}

\usepackage{microtype}

\usepackage[]{footmisc}
\usepackage{amsmath}
\usepackage{inconsolata}
\usepackage{cleveref}
\crefformat{section}{\S#2#1#3}
\crefformat{subsection}{\S#2#1#3}
\crefformat{subsubsection}{\S#2#1#3}
\crefrangeformat{section}{\S\S#3#1#4 to~#5#2#6}
\crefmultiformat{section}{\S\S#2#1#3}{ and~#2#1#3}{, #2#1#3}{ and~#2#1#3}

\Crefformat{figure}{#2Fig.~#1#3}
\Crefmultiformat{figure}{Figs.~#2#1#3}{ and~#2#1#3}{, #2#1#3}{ and~#2#1#3}
\Crefformat{table}{#2Tab.~#1#3}
\Crefmultiformat{table}{Tabs.~#2#1#3}{ and~#2#1#3}{, #2#1#3}{ and~#2#1#3}
\Crefformat{appendix}{#2Appx.~\S#1#3}
\crefformat{algorithm}{Alg.~#2#1#3}
\usepackage{graphicx}
\usepackage{caption}
\usepackage{subcaption}
\usepackage{booktabs}
\usepackage{multirow}
\usepackage{xspace}
\usepackage{enumitem}
\graphicspath{{imgs}}
\usepackage{algorithm}
\usepackage{algpseudocode}
\usepackage[]{mdframed}

\usepackage{amssymb}
\usepackage{pifont}
\usepackage{longtable}

\definecolor{purple}{HTML}{CCC0F0}
\definecolor{green}{HTML}{C1EDC7}
\definecolor{light_orange}{HTML}{F6D9B0}
\definecolor{pink}{HTML}{F6B1B3}
\definecolor{lightblue}{HTML}{AEDCF4}
\newcommand\crule[3][black]{\textcolor{#1}{\rule{#2}{#3}}}
\definecolor{c_subword}{HTML}{FDB462}
\definecolor{c_character}{HTML}{FB8072}
\definecolor{c_general}{HTML}{8DD3C7}
\definecolor{c_math}{HTML}{FDB462}
\definecolor{c_theorem}{HTML}{FB8072}
\definecolor{c_code_open}{HTML}{BEBADA}
\definecolor{c_code_python}{HTML}{7570B3}
\definecolor{c_direct}{HTML}{FDB462}
\definecolor{c_open}{HTML}{FB8072}
\usepackage{xcolor}
\usepackage{pifont}
\usepackage[normalem]{ulem}
\usepackage{tcolorbox}  % add this to your preamble
\tcbset{
  colframe=black, 
  colback=gray!10, 
  boxrule=0.5mm, 
  arc=4mm,
  before skip=1pt,   % space before the box
  after skip=5pt     % space after the box (adjust this as needed)
}

\makeatletter
  \newcommand\reduline{\bgroup\markoverwith{\textcolor{red}{\rule[-0.5ex]{2pt}{0.4pt}}}\ULon}
  \UL@protected\def\reduwave{\leavevmode \bgroup 
    \ifdim \ULdepth=\maxdimen \ULdepth 3.5\p@
    \else \advance\ULdepth2\p@ 
    \fi \markoverwith{\lower\ULdepth\hbox{\textcolor{red}{\sixly \char58}}}\ULon}
\makeatother
\title{LLM The Genius Paradox: A Linguistic and Math Expert's Struggle with Simple Word-based Counting Problems}
  \author{Nan Xu, Xuezhe Ma\\
University of Southern California\\
\texttt{\{nanx,xuezhema\}@usc.edu}}

\begin{document}
\maketitle
\begin{abstract}
% Recently, large language models have become experts in diverse areas requiring strong capabilities.
% , e.g., complex reasoning and math problem-solving. 
Interestingly, LLMs yet struggle with some basic tasks that humans find trivial to handle, e.g., counting the number of character $r$'s in the word ``strawberry''. There are several popular conjectures (e.g., \emph{tokenization}, \emph{architecture} and \emph{training data}) regarding the reason for deficiency of LLMs in simple word-based counting problems, sharing the similar belief that such failure stems from model pretraining hence probably inevitable during deployment. 
% Considering the broader significance of counting tasks to general language tasks, 
In this paper, we carefully design multiple evaluation settings to investigate validity of prevalent conjectures. Meanwhile, we measure transferability of advanced mathematical and coding reasoning capabilities from specialized LLMs 
% (e.g., Qwen2 Math and CodeGemma) 
to simple counting tasks. Although specialized LLMs suffer from counting problems as well, we find conjectures about inherent deficiency of LLMs invalid and further seek opportunities to elicit knowledge and capabilities from LLMs that are beneficial to counting tasks. Compared with strategies such as \emph{finetuning} and \emph{in-context learning} that are commonly adopted to enhance performance on new or challenging tasks, we show that engaging \emph{reasoning} is the most robust and efficient way to help LLMs better perceive tasks with more accurate responses.

We hope our conjecture validation design could provide insights into the study of future critical failure modes of LLMs. Based on challenges in transferring  advanced capabilities to much simpler tasks, we call for more attention to model capability acquisition and evaluation. We also highlight the importance of cultivating consciousness of ``reasoning before responding'' during model pretraining.~\footnote{Codes and datasets are available at \url{https://github.com/xunannancy/LLMParadox.git}.}
\end{abstract}
\section{Introduction}
Recently, large language models (LLMs) are able to achieve human-level performance on tasks such as complex reasoning, taking proficiency exams, code generation, multilingual understanding, and math problem solving~\citep{llama2024,openai_reasoning}. They even obtain expert-level performance on more challenging tasks like Mathematical Olympiad~\citep{deepmind_imo_2024}.

Surprisingly, LLMs yet struggle with some basic tasks that are easy or trivial for humans to handle, where no extensive word knowledge or complicated reasoning is required~\citep{ball2024can,shin2024large,yehudai2024can}. For instance, GPT-4o generates a wrong answer to the questions of counting the number of character $r$'s in the word ``strawberry''~\citep{karpathy_tweet_2024_tokenization}.
%although even kids unaware of the word can easily tell there are three rather two by simply perceiving the surface pattern of the word. 

The research community has discussed actively over the mysterious reason for such unexpected failures. The top-voted conjecture attributes such deficiency in counting characters to the \emph{subword tokenization algorithm} adopted by prevalent LLMs~\citep{shin2024large,karpathy_tweet_2024_tokenization}. Other researchers speculate that LLMs haven't seen \emph{sufficient character-level data} during pretraining, hence lack the capability to understand character-level relationships~\citep{shin2024large}. 
\citet{yehudai2024can} theoretically proved that the capability of transformer-based models to count characters is \emph{constrained by their embedding size}, and the more unique characters in words the worse performance achieved by LLMs.
All the prior conjectures suggest that the deficiency of LLMs in solving easy word-based counting tasks originates from the design of LLM systems (i.e., tokenization or model size) or the pretraining procedure (i.e., lack of character-level training), hence \textbf{inevitable} during model deployment.

Considering the broader impacts of word-based counting tasks on important research areas such as morphological analysis~\citep{shin2024large}, we focus on investigating validity of above conjectures on LLM failures by carefully designing multiple evaluation settings:
% \begin{itemize}
% % ~\\\\1) 
%  \item \textit{Conjecture I:} By evaluating LLMs with character-level perturbation and explicit character tokenization rather than the default subword input on counting tasks, we do not observe noticeable performance improvement of LLMs, hence confirm the incorrectness of \emph{tokenization issues} conjecture firstly. 
% % ~\\\\2) 
% \item \textit{Conjecture II:}
% We select classification tasks (e.g., sentiment analysis) that LLMs are proficient in with natural input. With rarely seen character input instead, we observe performance well beyond random guess. This implies strong capability of LLMs in performing character-level reasoning, which denies the conjecture of \emph{lacking character-level training}.  
% % ~\\\\3) 
% \item \textit{Conjecture III:}
% We conduct comprehensive analysis on the influence of unique characters within queried words but observe no clear connection to LLM performance, suggesting the inaccuracy of the third conjecture regarding \emph{model size}.
% \end{itemize}
\begin{tcolorbox}
  \textbf{Conjecture I:} By evaluating LLMs with character-level perturbation and explicit character tokenization rather than the default subword input, we do not observe noticeable performance improvement of LLMs, hence invalidating the conjecture regarding \emph{subword tokenization}. 
\end{tcolorbox}
\begin{tcolorbox}
  \textbf{Conjecture II:} We consider classification tasks (e.g., sentiment analysis) that LLMs are proficient in with natural input. With input phrased by rarely seen character format instead, we observe performance well beyond random guess. This implies strong capability of LLMs in performing character-level reasoning, which conflicts with \emph{lack of character-level training} conjecture.  
\end{tcolorbox}
\begin{tcolorbox}
  \textbf{Conjecture III:} We conduct comprehensive analysis on the impact of unique characters within queried words, but observe no clear connection to LLM performance, implying invalidness of the conjecture regarding \emph{model size}.
\end{tcolorbox}
We further evaluate specialized models, such as Qwen2 Math~\citep{qwen2_math} and CodeGemma~\citep{team2024codegemma}, aiming to measure transferability of advanced mathematical and coding reasoning capability to simple word-based counting tasks. Unfortunately, neither math nor coding LLMs are able to improve performance over their base model trained on general domains in the open-ended setting.
In contrast, python codes explicitly requested from coding LLMs can complete counting tasks with perfection. Such undesired failure from powerful specialized LLMs trained on way more challenging task data calls for more research in training strategies for capability acquisition and benchmark construction for comprehensive capability evaluation.

Motivated by invalidation of prevalent conjectures regarding inherent deficiency of LLMs, we seek opportunities to elicit knowledge and reasoning capabilities from LLMs that are beneficial to simple counting tasks. Specifically, we evaluate effectiveness of strategies commonly used to enhance LLM performance on new or challenging tasks, i.e., widely adopted reasoning methods such as \emph{chain-of-thought}~\citep{wei2022chain} and \emph{self-consistency}~\citep{wang2022self}, finetuning on counting data as well as in-context learning (ICL)~\citep{brown2020language,wei2023larger}. Compared with the other two directions, we find engaging reasoning is the most robust and efficient way to help LLMs better perceive the task and enhance final performance. With the aid of reasoning, GPT-4o is able to address all studied counting tasks with perfection, which is consistent with the core idea of ``complex reasoning before responding'' underlying recently announced model OpenAI o1~\citep{openai_reasoning}.

In summary, we analyze existing conjectures over failure modes of LLMs on simple word-based counting problems. We hope our conjecture validation procedure could also give insightful guidance in studying other unsolved deficiencies of LLMs such as the lost-in-the-middle phenomenon~\citep{liu-etal-2024-lost}, distraction by irrelevant context~\citep{shi2023large,chen2024premise}, etc. We also show inability of specialized math or coding LLMs to transfer advanced capabilities to much simpler tasks, calling for more attention and research in model capability acquisition during training and comprehensive capability evaluation during benchmarking. Lastly, we find effectiveness of reasoning strategies to help elicit knowledge and problem-solving capabilities from LLMs, highlighting importance of cultivating consciousness of reasoning during model pretraining.

\section{Background}
% From behaviors of leading large language models, we have observed significant deficiency in handling simple word-based counting problems. These models are mostly equipped with a dense transformer~\citep{vaswani2017attention} architecture~\footnote{Other emergent architectures, such as Mamba~\citep{gu2023mamba} and Liquid Neural Network~\citep{hasani2021liquid}, are promising to handle sequences more efficiently than dominant transformer-based LLMs while achieve competitive performance (e.g., LFM~\citep{liquid_ai_2024} from Liquid AI, FalconMamba~\citep{falconmamba} from TII). Since they are still under active development, we keep discussion to transformer-based models in this paper.} and share similar inference pipeline. 
% This section briefly describes how LLMs respond to prompts mainly from the perspective of input tokenization and output modeling, followed by the discussion of their jagged intelligence towards AI tasks of different challenging levels. 
We introduce \textbf{related work} in~\Cref{sec:related_work}.
\subsection{Tokenization}~\label{sec:tokenization}
% To convert text data to numerical data for model processing, tokenizers are developed to map text data into individual tokens and then convert those tokens into numbers, so that we can build tensors out of them and feed them to the model. Accordingly, a vocabulary is defined by the total number of independent tokens, each token gets assigned an ID starting from $0$ and going up to the vocabulary size, and LLMs use these IDs to identify each token. 
\emph{Word}-based tokenization algorithms used in earlier non-transformer models such as Word2Vec~\citep{mikolov2013efficient}, FastText~\citep{bojanowski-etal-2017-enriching} and GloVe~\citep{pennington-etal-2014-glove}, split texts into words (probably with some extra rules) and find numerical representation for each of them. Words that are unseen in the training corpus or ignored due to limited vocabulary size are typically represented by an unknown token, hence models lose their sensible information. On the contrary, \emph{character}- and \emph{byte}-based tokenization algorithms lead to much smaller vocabularies and far fewer out-of-vocabulary tokens by splitting texts into characters (e.g., CharBERT~\citep{ma-etal-2020-charbert} and Char2Subword~\citep{aguilar-etal-2021-char2subword-extending}) and bytes (e.g., Canine~\citep{clark-etal-2022-canine} and Byt5~\citep{xue-etal-2022-byt5}), respectively. However, character- and byte-based token representations are less meaningful with sequence length drastically expanded, posing the challenging efficiency issue for modeling. 

To take advantage of both worlds, \emph{subword} tokenization algorithms, such as Byte-Pair Encoding (BPE)~\citep{sennrich2015neural}, WordPiece~\citep{wu2016google} and UnigramLM~\citep{kudo2018subword}, decompose rare words into meaningful subwords while keep frequently used words intact. With the aid of open-source fast tokenization tool tiktoken~\citep{tiktoken_github}, BPE has become the dominant tokenization algorithm adopted by recent large language models like Llama 3~\citep{dubey2024llama} and GPT-4o~\citep{openai_gpt4o}.
We visualize how tokenizers utilized by LLMs split the text differently, \emph{How many r's in the word "strawberry"}, as follows.
~\\\textbf{GPT-4o}: \texttt{\colorbox{purple}{How}\colorbox{green}{ many}\colorbox{light_orange}{ r}\colorbox{pink}{'s}\colorbox{lightblue}{ in}\colorbox{purple}{ the}\colorbox{green}{ word}}
\texttt{\colorbox{light_orange}{ "}\colorbox{pink}{\reduwave{st}}\colorbox{lightblue}{\reduwave{raw}}\colorbox{purple}{\reduwave{berry}}\colorbox{green}{"?}}
~\\\textbf{Llama 3}: \texttt{\colorbox{purple}{How}\colorbox{green}{ many}\colorbox{light_orange}{ r}\colorbox{pink}{'s}\colorbox{lightblue}{ in}\colorbox{purple}{ the}\colorbox{green}{ word}}
\texttt{\colorbox{light_orange}{ "}\colorbox{pink}{\reduwave{str}}\colorbox{lightblue}{\reduwave{aw}}\colorbox{purple}{\reduwave{berry}}\colorbox{green}{"?}}
~\\\textbf{Gemma 1}: \texttt{\colorbox{purple}{How}\colorbox{green}{ many}\colorbox{light_orange}{ r}\colorbox{pink}{'}\colorbox{lightblue}{s}\colorbox{purple}{ in}\colorbox{green}{ the}}
\texttt{\colorbox{light_orange}{ word}\colorbox{pink}{ "}\colorbox{lightblue}{\reduwave{strawberry}}\colorbox{purple}{"?}}
~\\\textbf{Mistral v0.3}: \texttt{\colorbox{purple}{How}\colorbox{green}{ many}\colorbox{light_orange}{ r}\colorbox{pink}{'}\colorbox{lightblue}{s}\colorbox{purple}{ in}\colorbox{green}{ the}
\texttt{\colorbox{light_orange}{ word}\colorbox{pink}{ "}}\colorbox{lightblue}{\reduwave{st}}\colorbox{purple}{\reduwave{raw}}\colorbox{green}{\reduwave{berry}}\colorbox{light_orange}{"?}}
~\\\textbf{DeepSeek V2}: \texttt{\colorbox{purple}{How}\colorbox{green}{ many}\colorbox{light_orange}{ r}\colorbox{pink}{'}\colorbox{lightblue}{s}\colorbox{purple}{ in}\colorbox{green}{ the}
\texttt{\colorbox{light_orange}{ word}\colorbox{pink}{ "}}\colorbox{lightblue}{\reduwave{straw}}\colorbox{purple}{\reduwave{berry}}\colorbox{green}{"?}}
~\\\textbf{Yi 1.5}: \texttt{\colorbox{purple}{How}\colorbox{green}{ many}\colorbox{light_orange}{ r}\colorbox{pink}{'}\colorbox{lightblue}{s}\colorbox{purple}{ in}\colorbox{green}{ the}\colorbox{light_orange}{ word}
\texttt{\colorbox{pink}{ "}}\colorbox{lightblue}{\reduwave{st}}\colorbox{purple}{\reduwave{raw}}\colorbox{green}{\reduwave{berry}}\colorbox{light_orange}{"}\colorbox{pink}{?}}
% ~\\As shown above, the word \emph{strawberry} is less frequently used compared with other words such as \emph{How} and \emph{many}, hence decomposed into more common subwords, e.g., 
% % \texttt{\colorbox{purple}{st}\colorbox{green}{raw}\colorbox{purple}{light_orange}} 
% \texttt{st}, \texttt{raw}, and \texttt{berry},
% by the tokenizers of GPT-4o, Mistral and Yi. 
\subsection{Language Modeling}
Given a sequence of $m$ discrete tokens $C=\{x_1,\dots,x_{m}\}$ decomposed by the tokenizer, the language model predicts the next token according to the learned distribution $P_{\theta}$ parameterized by $\theta$. Following different decoding strategies, the model generates $n$ more tokens step-by-step:
\begin{equation}
p(x_{m+1:m+n}|\mathcal{C})=\prod^n_{t=1}P_{\theta}(x_{t}|\mathcal{C},x_{m+1}\dots x_{m+t-1}).\nonumber%\label{eq:prob}
\end{equation}

When the context $C$ represents a question from the user, the continuation $\{x_{m+1},\dots,x_{m+n}\}$ from the instructed or chat model can be 1) the direct answer, 2) reasoning process followed by the final answer~\citep{wei2022chain,kojima2022large}, or 3) the final answer followed by detailed explanation~\citep{xie2024order}.
\section{Experimental Setup}~\label{sec:exp_setup}
\begin{table*}[t!]
\centering
\resizebox{\textwidth}{!}{%
\begin{tabular}{@{}lccccccccc@{}}
\toprule
Task                & \bf GPT-4o & \bf Llama 3 & \bf Qwen 1.5 & \bf Gemma 1 & \bf InternLM 2 & \bf Phi 3 & \bf Mistral v0.3 & \bf DeepSeek V2 & \bf Yi 1.5 \\ \midrule
\bf I: Char Occur       & 82.4   & 34.6    & 30.6     & 41.2    & 60.8      & 39.0  & 35.4         & 27.2        & 46.6   \\
\bf II: Substring Occur & \bf 87.4   & 58.2    & 58.9     & 50.8    & 50.3      & 73.0  & 57.7         & 61.5        & 59.6   \\
\bf III: Word Len       & \bf 92.0   & 74.6    & 42.4     & 26.0    & 55.8      & 64.4  & 41.0         & 36.6        & 58.0   \\
\bf IV: Distinct Char   & \bf 89.2   & 57.8    & 27.8     & 4.4     & 21.8      & 70.2  & 34.4         & 28.4        & 36.6   \\\midrule 
\textbf{MMLU} (0-shot) &85.0&64.2&58.5&50.4&59.2&75.7&59.3&53.0&67.0\\
\textbf{GSM8K} (0-shot) &86.3&78.9&58.0&38.6&67.9&82.2&47.8&70.6&81.8\\
\bottomrule
\end{tabular}%
}
\caption{Performance of LLMs on simple word-based counting problems, as well as on the general benchmark MMLU and the math benchmark GSM8K for comparison. Both open-source and proprietary LLMs struggle with answering the correct numbers (i.e., Task I, III and IV) or identifying existence of subtrings in words (i.e., Task II). After marking counting accuracy ($2$nd to $5$th rows) higher than general and math benchmarks (bottom two rows) in \textbf{boldface}, we find LLMs can hardly achieve much better performance than that on the more challenging benchmark MMLU and GSM8K.}
\label{tab:basic_performance}
        \vspace{-1em}
\end{table*}
% In this section, we describe the general experimental setup used to evaluate LLMs on word-based counting problems.
% \paragraph{Evaluation Benchmarks} 
Motivated by the problem of counting the number of r's in the word ``strawberry''~\citep{karpathy_tweet_2024_jagged}, we randomly sample $500$ words from the NLTK library~\citep{bird2009natural} and prompt LLMs to answer four distinct word-based questions in zero-shot listed as follows, with their statistics listed in~\Cref{tab:dataset_statistics}.
~\\\textbf{Task I (Char Occur):} \texttt{How many \{x\}'s in the word ``\{Y\}''?}

In this task, $x$ is a character randomly sampled from the word $Y$. For example, given the question ``How many r's in the word ``strawberry''?'', the correct answer should be $3$.
~\\\textbf{Task II (Substring Occur):} \texttt{Is the substring ``\{x\}'' part of the word ``\{Y\}''?}

In this task, $x$ is composed of a set of characters, and could be present or absent from the word $Y$. For instance, the answer to the question ``Is the substring ``raw'' part of the word ``strawberry''?'' is ``Yes'', while ``No'' is the answer to the question when the substring is substituted by ``rae''~\footnote{To mitigate potential bias of LLM towards affirmative or negative response, we randomly extract one substring from the word with the positive answer and replace one of the characters so that the answer switches to negative, resulting in one positive and one negative instance per word.}. 
~\\\textbf{Task III (Word Len):} \texttt{How many characters in the word ``Y''?}

This task requires LLMs to accurately count the number of characters in one word. For example, the ground-truth answer to the question ``How many characters in the word ``strawberry''?'' is $10$.
~\\\textbf{Task IV (Distinct Char):} \texttt{How many distinct characters in the word ``Y''?}

Different from Task III, the LLMs are examined whether they are able to recognize each character in the word as well as their frequency. For instance, given the question ``How many distinct characters in the word ``strawberry''?'', the correct answer is $8$ since $r$ repeats three times and should be considered one single character.

We provide detailed introduction to evaluated \textbf{language models} and \textbf{evaluation metrics} in~\Cref{sec:experimental_setup_appendix}. In~\Cref{tab:basic_performance}, we show evaluation results of different models on four studied tasks and two widely adopted benchmarks for comparison. Although the counting problems do not require extensive world knowledge or math problem-solving abilities, all studied LLMs struggle with these seemingly simple tasks, resulting in similar or even worse accuracy than that on MMLU and GSM8K. 

In this work, we mainly focus on the English domain, but observe similar issues (shown in~\Cref{tab:lan_performance}) when asked character occurrence questions in words from other Germanic (i.e., German and Swedish) and Romance (i.e., French, Spanish, Italian and Portuguese) Languages. 
% We leave deficiency of LLMs in other languages for future work.
We leave \emph{multilingual} analysis for future work. 

Besides static prompts used for four counting tasks, we find that \emph{prompt engineering} does not enhance llm capabilities to circumvent such failure modes. We study two types of prompts per task conveying similar semantic meanings to evaluate effectiveness of prompt engineering: 1) four paraphrased prompts created by human heuristics and 2) one improved prompt provided by Claude~\footnote{\url{https://www.anthropic.com/news/prompt-improver}}.
% ~\footnote{Prompt improver (~\url{https://www.anthropic.com/news/prompt-improver}) from Claude introduced the ability to improve prompts for better completions, which takes existing prompts and leverages Claude to automatically refine them using advanced prompt engineering techniques. NOTE that we have removed chain-of-thought reasoning instruction and examples in the improved prompts to form fair comparison with original simple prompts.}. 
We provide detailed analysis in~\Cref{sec:prompt_engineering}.

\begin{figure*}[t!]
     \centering
    \begin{subfigure}[b]{\textwidth}
         \centering
         \includegraphics[width=\linewidth]{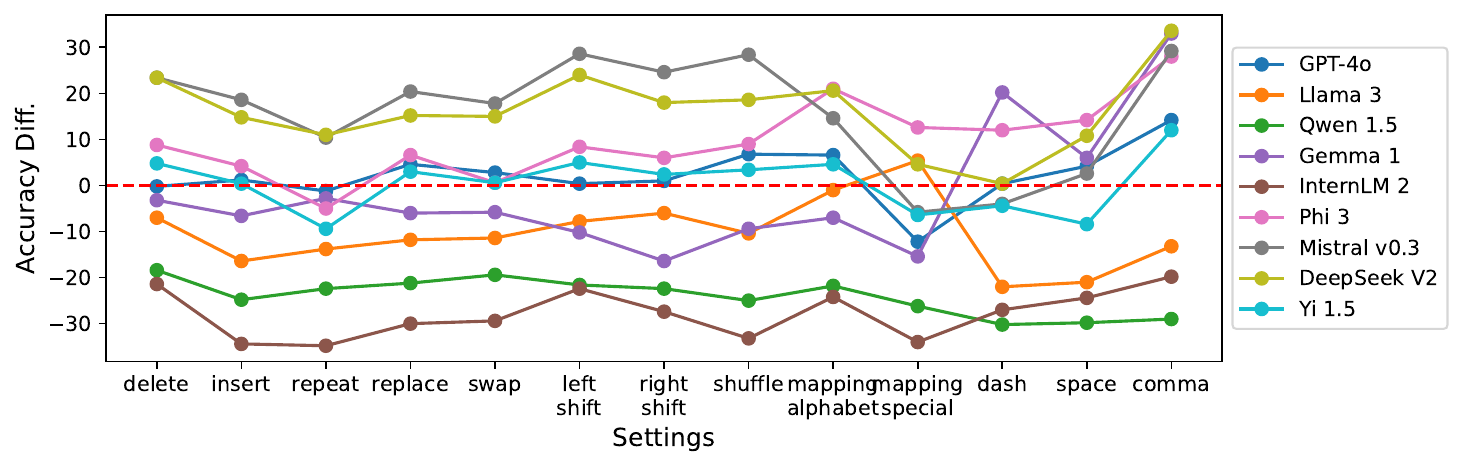}\caption{Performance difference after switching to  implicit character-level tokenization by adopting character-level word perturbations.}
         \label{fig:pereturbation}
     \end{subfigure}
         \begin{subfigure}[b]{\textwidth}
         \centering
         \includegraphics[width=\linewidth]{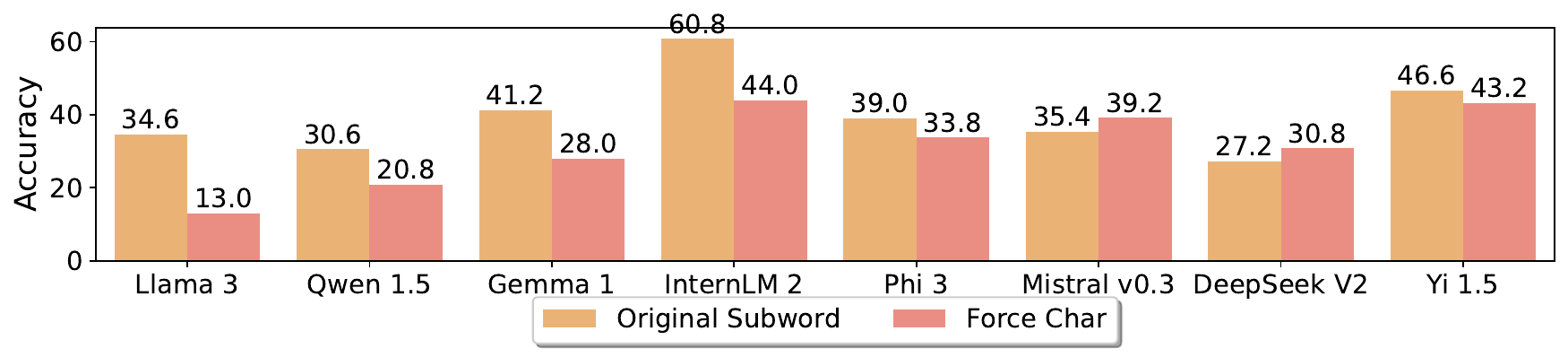}\caption{Performance comparison between original subword tokenization (\crule[c_subword]{.3cm}{.3cm}) and explicitly forced character tokenization (\crule[c_character]{.3cm}{.3cm}).}
         \label{fig:force_tokenization_strawberry_text}
     \end{subfigure}
        \caption{Performance comparison on \textbf{Task I} with different tokenization strategies to verify \textbf{Conjecture I} that deficiency in word-based counting is caused by subword tokenization of LLMs.\textbf{Top:} There is no perturbation that benefits all studied models identifying core character-level information (above $0$). \textbf{Bottom:} Directly feeding character tokens instead of subword tokens does not help LLMs perceive individual characters within words. Above comparison as well as results on other three studied tasks shown in~\Cref{fig:tokenization_extra} refuse the conjecture regarding LLM tokenization.}
        \label{fig:tokenization}
        \vspace{-1em}
\end{figure*}

\section{Why LLMs Struggle with Simple Counting Problems}\label{sec:why}

% Considering jagged intelligence towards simple word-based counting problems, we can summarize existing efforts to explain the failure mode of LLMs
There are three major conjectures trying to explain LLMs deficiency in simple word-based counting problems, detailed as follows:
~\\\textbf{Conjecture I: Tokenization Issues} 
~\\As introduced in~\Cref{sec:tokenization}, subword tokenization has become the dominant algorithm to convert text into numerical representations, making it challenging to perceive intrinsic characteristics and nuances of individual characters within words ~\citep{karpathy_tweet_2024_tokenization,shin2024large}. Moreover, the relationship between individual characters within a word can hardly be captured by the attention mechanism.
~\\\textbf{Conjecture II: Lacking Character-level Training} 
~\\Existing large language models are mainly pre-trained and post-trained on word-level data~\citep{bai2023qwen,cai2024internlm2,dubey2024llama}, hence not optimized for tasks that require character-level analysis.
~\\\textbf{Conjecture III: Excessive Unique Characters within Words} 
~\\Recent literature theoretically proves that the capability of transformers to count letters in words is upper bounded by their embedding size~\citep{yehudai2024can}. Empirically they find that Gemini~\citep{reid2024gemini} tends to make more mistakes when distinct characters in words increase.

In this section, we verify the above conjectures one-by-one by designing characteristic settings and comparing LLM performance with the default one introduced in~\Cref{sec:exp_setup}. 
\subsection{Conjecture I: Tokenization Issues}
To verify the conjecture ``LLMs fail on simple word-based count problems due to the subword tokenization'', we 
\emph{
~\\\null$\quad\bullet$ Design settings where the tokenizer has to implicitly or explicitly split texts into \textbf{characters} rather than subwords.
~\\\null$\quad\bullet$ Check the performance of LLMs on four counting tasks within the new tokenization.
~\\\null$\quad\bullet$ Analyze the implications: 
~\\\null$\quad1)$ If we observe noticeable improvement, then the conjecture is correct; 
~\\\null$\quad2)$ If the performance maintains similar or even degrades, then the conjecture is invalid.
}

\paragraph{Settings} To expose more character-level information of the studied word to LLMs, we follow literature that studies robustness of NLP models~\citep{liu2020joint,moradi-samwald-2021-evaluating,rocamorarevisiting} by conducting character-level word perturbations: 1) delete, 2) insert, 3) repeat, 4) replace, 5) swap, 6) left shift, 7) right shift, 8) shuffle, 9) mapping to alphabetical character, and 10) mapping to special character~\footnote{The answers may change after character-level perturbations like \emph{delete}, \emph{insert}, \emph{repeat}, and \emph{replace}, while keeping unaltered after perturbations such as \emph{swap}, \emph{left}/\emph{right shift}, \emph{shuffle} and \emph{alphabetical}/\emph{special mapping}.}. Besides, we also manually split characters among the word without altering the final answer by adding special characters in between: 1) dash, 2) space, and 3) comma. In~\Cref{tab:implicit_tokenization}, we present an example for every character-level word perturbation method.

Beyond above implicit character-level tokenization, we explicitly interfere with the tokenization process during inference so that the studied word is tokenized into a list of individual character tokens while other words are tokenized to subwords.
\paragraph{Results} In~\Cref{fig:tokenization}, we demonstrate performance comparison when LLMs are provided with the original subword tokens and our proposed implicit (top) or explicit (bottom) character tokens. Noticeably, LLMs do not benefit from inputs represented by either implicit or explicit character tokens to better perceive character-level information of key words, leading to similar or even worse performance than that given subword input. Therefore, we empirically \textbf{refute} the popular conjecture that the subword tokenization leads to LLMs failure in word-based counting tasks.
\subsection{Conjecture II: Lacking Character-level Training}
\begin{figure*}[t!]
     \centering
    % \begin{subfigure}[b]{\textwidth}
    %      \centering
    %      \includegraphics[width=\linewidth]{classification_emotion.pdf}\caption{Emotion.}
    %      \label{fig:emotion}
    %  \end{subfigure}
     %     \begin{subfigure}[b]{\textwidth}
     %     \centering
     %     \includegraphics[width=\linewidth]{classification_imdb.pdf}\caption{IMDB.}
     %     \label{fig:imdb}
     % \end{subfigure}
          \includegraphics[width=\linewidth]{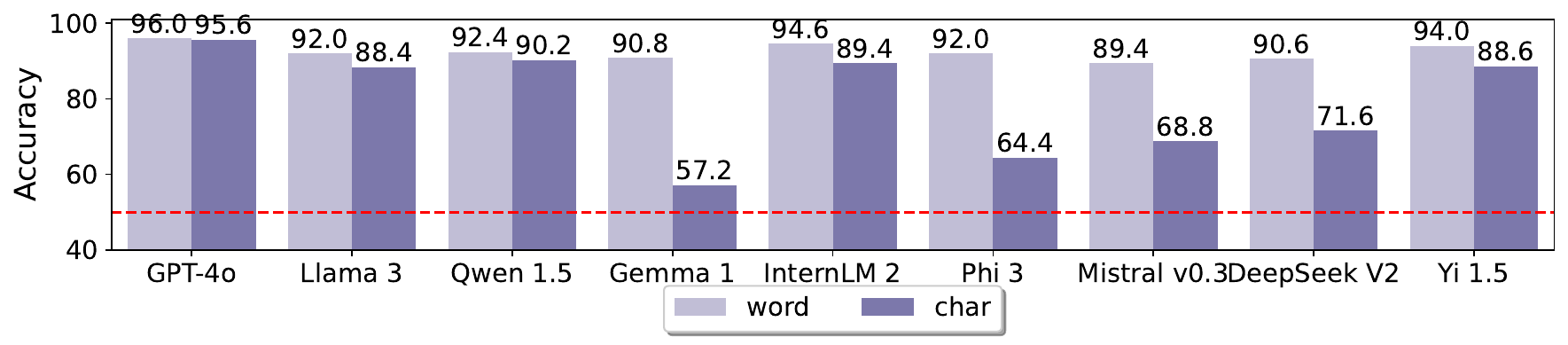}
     %     \begin{subfigure}[b]{\textwidth}
     %     \centering
     %     \includegraphics[width=\linewidth]{classification_sst2.pdf}\caption{SST-2.}
     %     \label{fig:sst2}
     % \end{subfigure}
        \caption{Performance comparison of LLMs between natural word and character input on classification dataset \emph{IMDB}. Although questions represented by individual characters are rarely seen, all studied models are able to achieve accuracy much higher than random guess denoted by \textcolor{red}{- - -}. The minor performance drop compared with natural word input implies that LLMs have the ability to handle tasks requiring character-level understanding, rejecting the conjecture relevant to lack of character training. Figure \ref{fig:classification_more} shows similar observations on other datasets.}
        \label{fig:classification}
        \vspace{-1em}
\end{figure*}
\begin{figure*}[t!]
     \centering
     \includegraphics[width=\linewidth]{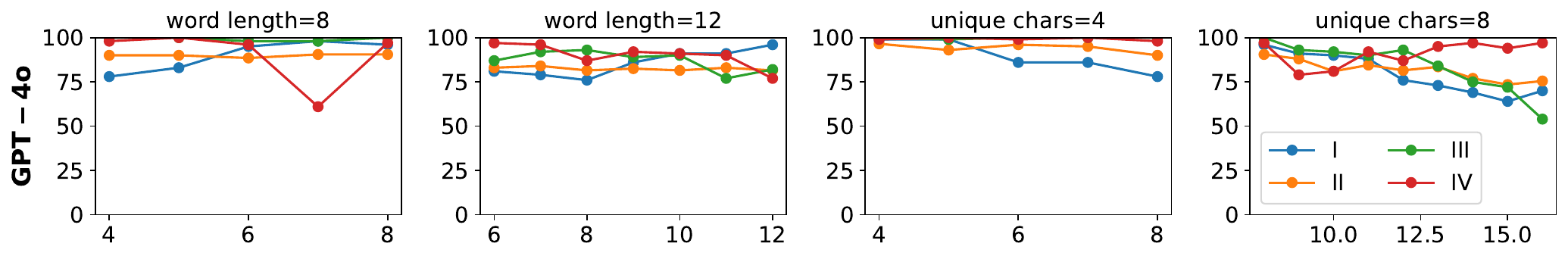}%\caption{GPT-4o.}
    % \begin{subfigure}[b]{\textwidth}
    %      \centering
    %      \includegraphics[width=\linewidth]{length_unique_gpt-4o.pdf}%\caption{GPT-4o.}
    %      \label{fig:unique_len_gpt_4o}
    %  \end{subfigure}
    %      \begin{subfigure}[b]{\textwidth}
    %      \centering
    %      \includegraphics[width=\linewidth]{length_unique_Meta-Llama-3-8B-Instruct.pdf}%\caption{Llama 3.}
    %      \label{fig:unique_len_llama3}
    %  \end{subfigure}
        \caption{Performance variation from GPT-4o on four tasks. \textbf{$1$st \& $2$nd column:} there is no noteworthy correlation between model performance and number of unique characters in queried words. \textbf{$3$rd \& $4$th column:} we observe clear trend of performance drop from both models when the word length keeps on increasing since 10. 
        We leave similar trend from Llama 3 in~\Cref{fig:unique_len_more}.
        }
        \label{fig:unique_len}
        \vspace{-1em}
\end{figure*}

In order to validate the correctness of the conjecture, ``LLMs haven't trained on sufficient character-level data, hence lack ability to understand and handle tasks requiring character-level reasoning,'' we
\emph{
~\\\null$\quad\bullet$ Evaluate performance of LLMs on tasks they are proficient in, but with character input.
~\\\null$\quad\bullet$ Compare the performance of LLMs between natural text and the rarely seen format of character input.
~\\\null$\quad\bullet$ Analyze the implications: 
~\\\null$\quad1)$ If we observe significant performance drop, then the conjecture is correct; 
~\\\null$\quad2)$ If the performance maintains similar or become slightly worse, then the conjecture is invalid.
}
\paragraph{Settings} We consider three sentiment analysis benchmarks where LLMs are able to achieve much higher accuracy than random guess in zero-shot setting: \emph{1) Emotion}: a dataset of $2000$ English Twitter messages with six basic emotions: anger, fear, joy, love, sadness, and surprise~\citep{saravia-etal-2018-carer}; \emph{2) IMDB}: a movie review dataset~\footnote{We randomly sample $500$ instances from the $25$k testing set for efficient inference. Note that we drop reviews containing more than $4000$ characters to ensure the input within the context length of all LLMs (i.e., $8192$).} for binary sentiment classification~\citep{maas-EtAl:2011:ACL-HLT2011}; \emph{3) SST-2}: $872$ single sentences~\footnote{We evaluate on the validation set since labels on the testing set are not publicly available.} extracted from movie reviews for binary classification~\citep{socher-etal-2013-recursive}. We present sentiment classification tasks as multiple-choice questions to LLMs, with options randomly ordered per question to avoid model bias towards specific options.
\paragraph{Results}
We demonstrate performance comparison between natural word and character input in~\Cref{fig:classification}. Without further tuning, all studied LLMs can perform sentiment analysis with accuracy above $90\%$ on binary classification and above $50\%$ on 6-way classification. Meanwhile, we observe minor performance drop when input format switches from natural words to rarely seen characters, which is still well above random guess performance. This suggests that pretrained LLMs have the capability to perform character-level reasoning, although similar data is not sufficiently seen during model pretraining or fine-tuning. Therefore, we \textbf{deny} the conjecture that deficiency of LLMs in simple word-based counting tasks is attributed to lack of training on similar data.
\subsection{Conjecture III: Excessive Unique Characters within Words}
\begin{figure*}[t!]
     \centering
    % \begin{subfigure}[b]{\textwidth}
    %      \centering
    %      \includegraphics[width=\linewidth]{math_code_strawberry_text.pdf}\caption{Task I: Char Occur.}
    %      \label{fig:math_code_strawberry_text}
    %  \end{subfigure}
         % \begin{subfigure}[b]{\textwidth}
         % \centering
         \includegraphics[width=\linewidth]{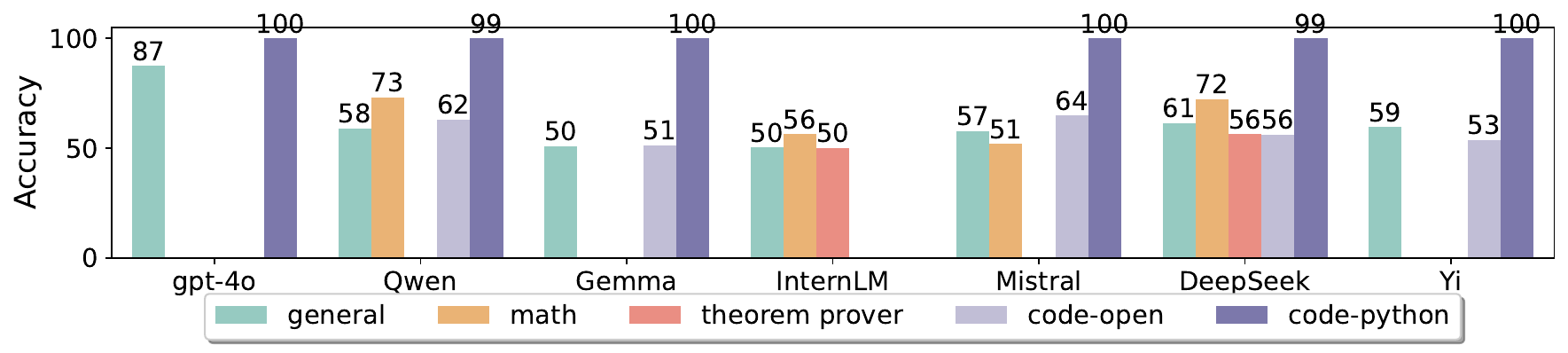}%caption{Task II: Substring Occur.}
     %     \label{fig:math_code_strawberry_substring_occurrence}
     % \end{subfigure}
        \caption{Performance comparison among LLMs trained on general and special domain data for \emph{Task II}. Models specialized in mathematical reasoning (\crule[c_math]{.3cm}{.3cm}) or theorem proving (\crule[c_theorem]{.3cm}{.3cm}) can not better handle word-based counting tasks than general (\crule[c_general]{.3cm}{.3cm}) ones. Although code models are able to write Python codes (\crule[c_code_python]{.3cm}{.3cm}) and solve tasks successfully, they fail to reason and answer accurately in open-ended setting (\crule[c_code_open]{.3cm}{.3cm}) . We leave results containing similar observations on three other tasks in~\Cref{fig:math_code_more}.}
        \label{fig:math_code}
        \vspace{-1em}
\end{figure*}

\citet{yehudai2024can} regard the counting problem as a more difficult one compared with the popular ``needle in haystack''~\citep{LLMTest_NeedleInAHaystack,ivgi-etal-2023-efficient}, since the former requires considering multiple occurrences of a given string, while the latter aims to retrieve only one appearance in a long context. They further find that \emph{the more unique characters showing up in the string, the more challenging for transformer-based LLMs to count the occurrence}. On the contrary, \emph{model performance is barely sensitive to the extension of string length}. We conduct systematic evaluation of model capabilities to handle word-based counting tasks when character uniqueness and total counts vary. 
\paragraph{Settings}
We select the closed-source model GPT-4o and open-source model Llama 3, then evaluate on four sets of words by keeping the total number of characters or the number of distinct characters fixed while varying the other: 1) $500$ words with $8$ characters, 2) $500$ words with $12$ characters, 3) $500$ words with $4$ distinct characters, and 4) $500$ words with $8$ distinct characters.  
\paragraph{Results} As shown in~\Cref{fig:unique_len}, opposite to observations discovered in prior work~\citep{yehudai2024can}, the increasing number of distinct characters in queried words does not lead to degraded performance on word-based counting problems. Instead, when the total number of characters reaches 10 and keeps increasing, we find obvious accuracy drop in both models. Hence, the conjecture that excessive unique characters in queried words lead to poor word-based counting performance is \textbf{incorrect}.
\section{Whether Math/Code Train Data Helps}
Recently, many open-resource base LLMs have been further tuned on billions or trillions math~\citep{qwen2_math,ying2024internlm,mistral_mathstral,shao2024deepseekmath} or formal theorem-proving data~\citep{wu2024leangithubcompilinggithublean,xin2024deepseek} in order to solve advanced mathematical problems that require complex, multi-step logical reasoning. Similarly, quite a few code models~\citep{codeqwen1.5,team2024codegemma,mistral_codestral,zhu2024deepseek,01ai_llm_for_code} have been built on top of base LLMs and additionally trained on diverse programming language datasets, demonstrating significant advancements in various aspects of code-related tasks such as code generation~\citep{chen2021evaluating,austin2021program}, completion~\cite{liu2023repobench} and insertion~\citep{allal2023santacoder}.

In this section, we focus on evaluating whether additional training on mathematical or coding data helps LLMs understand and improve reasoning over word-based counting tasks.

\paragraph{Results}
We provide detailed introduction to \textbf{evaluated models} and \textbf{implementation details} in~\Cref{sec:whether_setup_more}. We visualize performance of models with different capabilities in~\Cref{fig:math_code}. We observe that models additional trained on mathematical reasoning  can not bring obvious improvement over those trained on general-domain data. This indicates that their acquired reasoning capability over math problems is not sufficient to handle word-based counting tasks. On the other side, code models are able to solve the counting tasks successfully when prompted to generate Python codes explicitly, suggesting that the studied tasks are of easy level. Interestingly, the powerful code models fail when prompted in open-ended setting, implying that they do not distill problem-solving capabilities during training on code-specific tasks.

Although specialized LLMs substantially enhance coding or mathematical reasoning capabilities over general LLMs, they still struggle in solving easy word-based counting problems that require easy-level reasoning.

\section{How to Make LLMs Experts Again}
\begin{figure*}[t!]
     \centering
     \includegraphics[width=\linewidth]{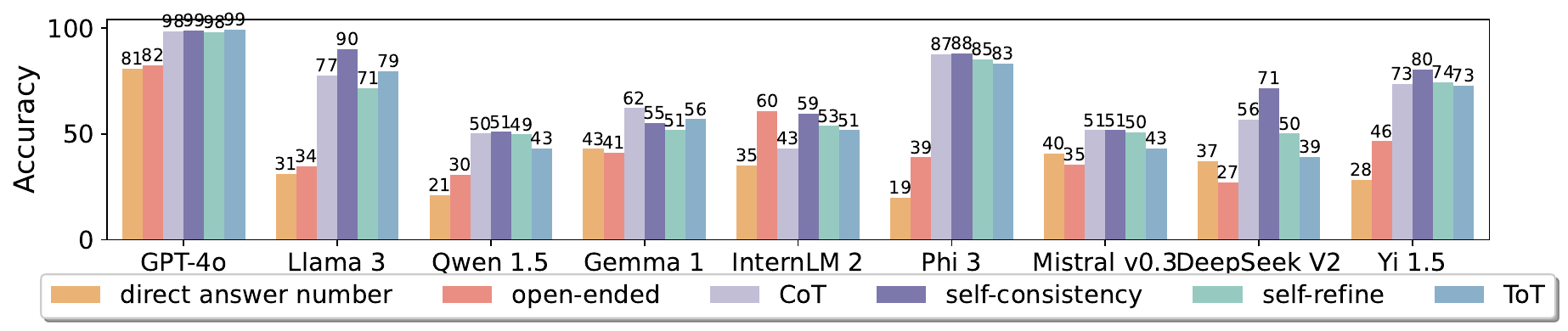}
        \caption{Benefits of applying different reasoning strategies to LLMs for \emph{Task I Char Occur}. We observe noticeable improvement from all studied reasoning strategies over baselines that particularly request numeric responses (\crule[c_direct]{.3cm}{.3cm}) or open-ended answers (\crule[c_open]{.3cm}{.3cm}). GPT-4o with the additional reasoning procedure can even solve tasks with perfection. We show similar improvement on other three tasks in~\Cref{fig:reasoning_others}.}
        \label{fig:reasoning_char_occur}
        \vspace{-1em}
\end{figure*}
\begin{table*}[t!]
\centering
\resizebox{\textwidth}{!}{%
\begin{tabular}{@{}lcccc|cccccc@{}}\toprule
\bf Finetune on & \bf Task I         & \bf Task II        & \bf Task III       & \bf Task IV         & \bf MMLU          & \bf IFEval        & \bf GPQA            & \bf Hellaswag       & \bf GSM8K         & \bf HumanEval       \\\midrule
-          & 34.6           & 58.2           & 74.6           & 57.8            & \textbf{64.3} & \textbf{68.6} & \textbf{30.6}   & 67.6            & \textbf{78.9} & 59.1            \\\midrule
\bf Task I     & \underline{\textbf{70.4\textcolor{blue}{↑}}} & 54.9\textcolor{red}{↓}          & 64.4\textcolor{red}{↓}          & 35.0\textcolor{red}{↓}           & 62.8\textcolor{red}{↓}         & 64.5\textcolor{red}{↓}         & 28.9\textcolor{red}{↓}           & \textbf{69.6 \textcolor{blue}{↑}} & 77.9\textcolor{red}{↓}         & \textbf{60.4 \textcolor{blue}{↑}} \\\midrule
\bf Task II    & 30.6\textcolor{red}{↓}          & \underline{\textbf{58.3\textcolor{blue}{↑}}} & \textbf{76.2\textcolor{blue}{↑}} & 31.4\textcolor{red}{↓}           & 62.4\textcolor{red}{↓}         & 59.3\textcolor{red}{↓}         & \textbf{30.6 \textcolor{blue}{↑}} & 59.0\textcolor{red}{↓}           & 76.6\textcolor{red}{↓}         & 59.1            \\\midrule
\bf Task III   & 28.2\textcolor{red}{↓}          & 56.6\textcolor{red}{↓}          & \underline{74.2\textcolor{red}{↓}}          & 22.8\textcolor{red}{↓}           & 63.5\textcolor{red}{↓}         & 62.8\textcolor{red}{↓}         & 27.7\textcolor{red}{↓}           & 53.8\textcolor{red}{↓}           & 77.1\textcolor{red}{↓}         & 57.3\textcolor{red}{↓}           \\\midrule
\bf Task IV    & 45.0\textcolor{blue}{↑}          & 56.4\textcolor{red}{↓}          & 39.4\textcolor{red}{↓}          & \underline{\textbf{87.2 \textcolor{blue}{↑}}} & 60.7\textcolor{red}{↓}         & 59.9\textcolor{red}{↓}         & 28.2\textcolor{red}{↓}           & 54.2\textcolor{red}{↓}           & 72.9\textcolor{red}{↓}         & 57.9\textcolor{red}{↓}\\\bottomrule   
\end{tabular}%
}
\caption{Performance of finetuned Llama 3 models on in- and out-distribution testing set. In each row, we train on task-specific data and test on same-distribution (results marked by \underline{\quad}) and other testing data, with performance improvement marked by \textcolor{blue}{↑} and drop by \textcolor{red}{↓} compared with the untuned Llama 3 model (- in $2$nd row). Finetuning on task-specific data does not necessarily enhance model capabilities in that task, even leading to worse performance in out-of-distribution tasks (\textbf{left} block) as well as other general, reasoning, math or coding benchmarks (\textbf{right} block).}
\label{tab:finetune_results}
        \vspace{-1em}
\end{table*}

As we have verified in~\Cref{sec:why}, the popular conjectures, such as tokenization and lack of character-level training, are not the true barriers for LLMs to solve counting tasks. Meanwhile, LLMs achieve competitive performance on far more challenging reasoning~\citep{clark2018think,zellers2019hellaswag,rein2023gpqa} and mathematical~\citep{cobbe2021training,hendrycks2021measuring} benchmarks. Therefore, we believe LLMs possess the knowledge and skills to solve counting problems if guided properly. We investigate whether reasoning strategies~\citep{wei2022chain,wang2022self,madaan2024self,sprague2024cot,yao2024tree} could elicit strong capabilities from LLMs to help perceive, reason and finally solve the problem.
% \subsection{Setup}\label{sec:reasoning_setup}
% We evaluate the ability of following methods to improve performance over directly prompting LLMs.
\paragraph{Reasoning Strategies} We investigate the following reasoning methods that have demonstrated great improvement in math and reasoning~\citep{sprague2024cot}: \emph{1) CoT}: chain-of-thought~\citep{wei2022chain} encourages models to reason before providing the final answer, which becomes the de facto method for eliciting reasoning capabilities from LLMs. \emph{2) self-consistency}: first samples a diverse set of reasoning paths instead of only taking the greedy one, and then selects the most consistent answer by majority voting~\citep{wang2022self}. \emph{3) self-refine}: uses a single LLM as the generator, refiner, and feedback provider~\citep{madaan2024self}. \emph{4) ToT}: tree-of-thought actively maintains a tree of thoughts, where each thought is a coherent language sequence that serves as an intermediate step toward problem solving~\citep{yao2024tree}. In contrast, we also append the instruction, ``\texttt{Directly answer the number}'' after each question, to request direct numeric answers from LLMs.
\begin{figure*}[t!]
     \centering
         \begin{subfigure}[b]{\textwidth}
         \centering
         \includegraphics[width=\linewidth]{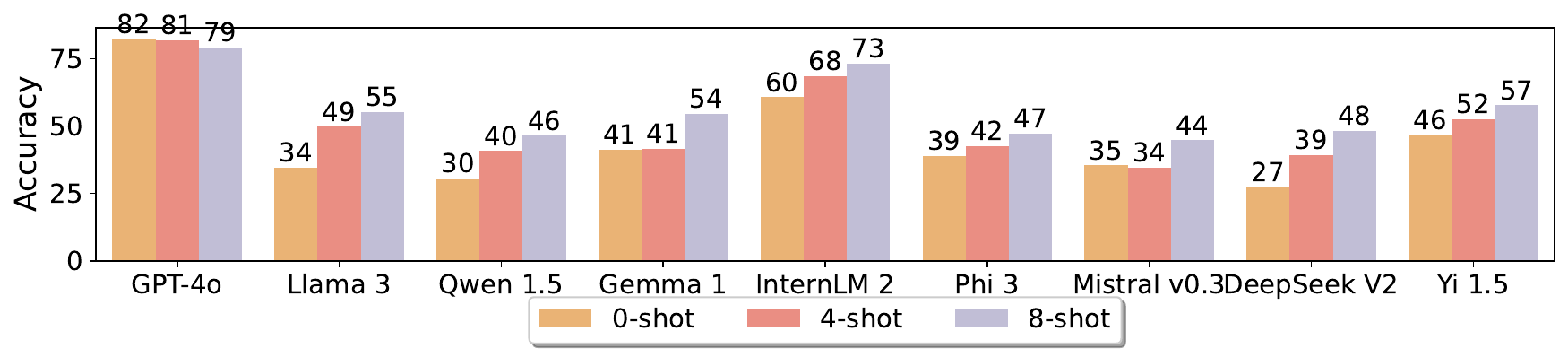}\caption{Task I: Char Occur.}
         \label{fig:icl_strawberry_char_occurrence}
     \end{subfigure}
    % \begin{subfigure}[b]{\textwidth}
    %      \centering
    %      \includegraphics[width=\linewidth]{icl_strawberry_substring_occurrence.pdf}\caption{Task II: Substring Occur.}
    %      \label{fig:icl_strawberry_substring_occurrence}
    %  \end{subfigure}
     %          \begin{subfigure}[b]{\textwidth}
     %     \centering
     %     \includegraphics[width=\linewidth]{icl_strawberry_word_length.pdf}\caption{Task III: Word Len.}
     %     \label{fig:icl_strawberry_word_length}
     % \end{subfigure}
         \begin{subfigure}[b]{\textwidth}
         \centering
         \includegraphics[width=\linewidth]{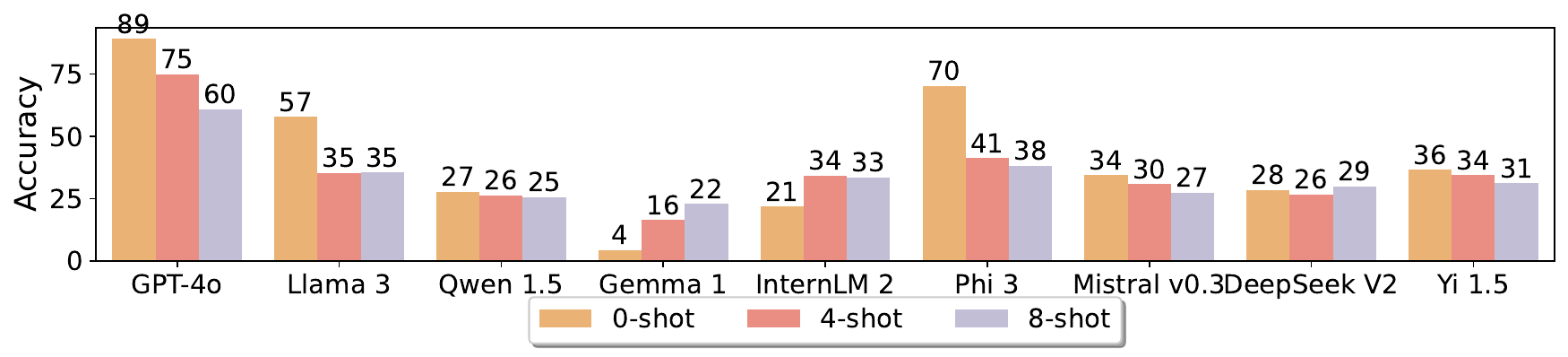}\caption{Task IV: Distinct Char.}
         \label{fig:icl_strawberry_distinct_characters}
     \end{subfigure}
        \caption{In-context learning performance of LLMs. Providing similar examples as demonstrations helps slightly improve performance of open-source models on Task I, while lead performance drop for most models on Task IV. We leave results on Task II and III in~\Cref{fig:icl_remains}.}
        \label{fig:icl}
        \vspace{-1em}
\end{figure*}
\paragraph{Other Strategies}
In supervised finetuning (SFT), collecting and mixing instruction tuning data are important steps to improve performance for specific capabilities~\citep{dubey2024llama}. Hence we \textbf{finetune} open-source LLMs with task-specific train data~\footnote{Motivated by benefits of reasoning procedures to counting tasks demonstrated in~\Cref{sec:reasoning_results}, we provide detailed reasoning before correct answers in the ground-truth responses.} and evaluate on both in-distribution test data and widely adopted benchmarks. By providing similar examples as context, \textbf{in-context learning} (ICL)~\citep{brown2020language,wei2023larger} has become another popular train-free method to efficiently improve LLM performance. We describe \textbf{implementation details} in~\Cref{sec:expert_implementation}.
\subsection{Reasoning}~\label{sec:reasoning_results}
In~\Cref{fig:reasoning_char_occur}, we compare diverse reasoning strategies introduced before with baseline strategies, i.e., directly responding with numeric values and open-ended generation. We find that all studied reasoning approaches are helpful to greatly improve performance over those without reasoning across four counting tasks, among which \emph{self-consistency} exhibits consistent advantage over other reasoning strategies for diverse LLMs. In addition, we show \textbf{scaling law} of \emph{self-consistency} in~\Cref{fig:scaling}, where no clear trend of performance boost as utilizing more reasoning paths is observed~\footnote{The observation is quite different from that shown in~\citep{wang2022self}, where using $40$ reasoning paths achieved the best performance by LaMDA-137B and GPT-3 code-davinci-001. We speculate that the model scale and task difficulty level may be the major reason.}. We provide case study from baseline strategies and CoT  in~\Cref{tab:case_study_strawberry_text}. 

With the aid of reasoning procedures, the most powerful model GPT-4o is capable of solving counting tasks with accuracy approaching $100\%$, indicating that the model can leverage its possessed knowledge and problem-solving abilities individually without external assistance. We also notice considerable performance margin from some LLMs between directly answering numerical value and open-ended generation, implying that they consciously invoke the reasoning process before providing the final answer for certain instances. We expect performance improvement in future LLMs if reasoning-related training is strengthened.
% and consciousness of reasoning necessity is enhanced. 

% \nan{add analysis regarding sensitivity to query phrase}
% \nan{add some case study as well}

\subsection{Supervised Finetuning}
In~\Cref{tab:finetune_results}, we evaluate capabilities of Llama 3 models finetuned on different task data. When training and testing on the same distribution, we do observe significant accuracy boost in \emph{Task I} (from $34.6\%$ to $70.4\%$) and \emph{Task IV} (from $57.8\%$ to $87.2\%$), while minor performance drop on the other two tasks. However, the acquired specific counting capability from training on \emph{Task I} or \emph{IV} can hardly transfer to other evaluated counting tasks, resulting in even much worse performance than the untuned model. Moreover, we find undesired lowered accuracy on benchmarks evaluating important capabilities such as reasoning and math, manifesting negative impacts of solely training on specific domains without considering other aspects. This emphasizes the importance of careful design for the proportion of different data sources, which is consistent with discoveries in literature~\citep{bai2023qwen,dubey2024llama} and leaves the finetuning strategy a less efficient way to improve performance of LLMs on new or challenging tasks compared with reasoning.

\subsection{In-context Learning}

We demonstrate the influence of demonstrations on counting tasks in~\Cref{fig:icl}. For \emph{Task I}, open-source LLMs achieve much higher accuracy in few-shot settings than zero-shot one, and more demonstrations exhibit further performance improvement. However, benefits of demonstrations are not always guaranteed. For example, additional example context greatly hurts performance of GPT-4o and the majority of open-source LLMs for \emph{Task II} (in~\Cref{fig:icl_strawberry_substring_occurrence}) and \emph{IV} (in~\Cref{fig:icl_strawberry_distinct_characters}).

\section{Conclusions}
% In this paper, we focus on the word-based counting tasks, which humans find trivial to handle but LLMs struggle with. 
By carefully designing multiple evaluation settings, we first show that prevalent conjectures regarding such unexpected failures are invalid.
% , suggesting possibility to enhance model capabilities to solve counting tasks. 
We further show that specialized models with advanced mathematical or coding reasoning capabilities also suffer from addressing simple counting problems. 
% Lastly, we investigate popular strategies that have exhibited effectiveness to enhance model performance on new or challenging tasks.
% , i.e., \emph{reasoning}, \emph{finetuning} and \emph{in-context learning}. 
We also find that \emph{reasoning} is the most robust and efficient way to aid models in better perceiving and solving tasks, highlighting more research into ``reasoning before responding'' during pretraining.
\clearpage
\section*{Limitations}
We investigate deficiency of diverse open-source LLMs as well as GPT-4o to address word-based counting problems. This work may have the following limitations: \emph{1) Lack of analysis on more proprietary LLMs:} for the sake of cost, we only consider GPT-4o and use it as the representative of other models of similar strong capabilities. Some online discussion has revealed similar issues from closed-source models such as Claude and Gemini. We hope researchers who develop these proprietary models can get insights from our conjecture validation procedure and reasoning-driven solutions, hence further boosting capabilities of top LLMs. \emph{2) Reasoning incorporated in pretraining:} we find that reasoning before providing the final answer during inference is effective in solving counting problems, while leaving training design of incorporating reasoning into pretraining as future direction.
\section*{Ethics Statement}
This paper presents comprehensive study of LLMs from diverse families that have gone through ethical reviews in prior works. Therefore, we believe our work does not pose additional ethical issues.
% \clearpage
% \bibliographystyle{acl_natbib}
\bibliography{custom}
\clearpage

\appendix
\section{Appendix}
\label{sec:appendix}
\begin{table*}[t!]
\centering
\resizebox{\textwidth}{!}{%
\begin{tabular}{@{}ll@{}}
\toprule
\bf Source                 & \bf Prompt                                                                                                                                                                                                                                                                                                                                                                                                                                                                                                                                                                                                                                                        \\ \midrule
\multicolumn{2}{c}{\bf Task I: Char Occur}                                                                                                                                                                                                                                                                                                                                                                                                                                                                                                                                                                                                                                                 \\
\bf Original Prompt        & How many \{\{char\}\}’s in the word “\{\{word\}\}”?                                                                                                                                                                                                                                                                                                                                                                                                                                                                                                                                                                                                           \\\midrule
\bf Human Prompt I         & How many times does the letter '\{\{char\}\}’' appear in the word "\{\{word\}\}"?                                                                                                                                                                                                                                                                                                                                                                                                                                                                                                                                                                             \\
\bf Human Prompt II        & Can you count the number of '\{\{char\}\}'s in "\{\{word\}\}"?                                                                                                                                                                                                                                                                                                                                                                                                                                                                                                                                                                                                \\
\bf Human Prompt III       & How many \{\{char\}\}'s can be found in "\{\{word\}\}"?                                                                                                                                                                                                                                                                                                                                                                                                                                                                                                                                                                                                       \\
\bf Human Prompt IV        & How often does the letter '\{\{char\}\}' show up in the word "\{\{word\}\}"?                                                                                                                                                                                                                                                                                                                                                                                                                                                                                                                                                                                  \\\midrule
\bf Claude Improved Prompt & \begin{tabular}[c]{@{}l@{}}You are tasked with counting the occurrences of a specific letter in a given word. Here are the inputs:\\ \\ Letter to count: \\ \textless{}letter\textgreater\\ \{\{letter\}\}\\ \textless{}/letter\textgreater\\ \\ Word to analyze:\\ \textless{}word\textgreater\\ \{\{word\}\}\\ \textless{}/word\textgreater\\ \\ Your task is to count how many times the specified letter appears in the given word. Consider the following:\\ - The letter is case-sensitive.\\ - Spaces and punctuation marks, if any, should be ignored.\\ \\ Please provide your response in a clear, concise sentence stating the count.\end{tabular} \\\midrule\midrule
\multicolumn{2}{c}{\bf Task II: Substring Occur}                                                                                                                                                                                                                                                                                                                                                                                                                                                                                                                                                                                                                                           \\
\bf Original Prompt        & Is the substring “\{\{substring\}\}” part of the word “\{\{word\}\}”?                                                                                                                                                                                                                                                                                                                                                                                                                                                                                                                                                                                         \\\midrule
\bf Human Prompt I         & Does the word "\{\{word\}\}" contain the substring "\{\{substring\}\}"?                                                                                                                                                                                                                                                                                                                                                                                                                                                                                                                                                                                       \\
\bf Human Prompt II        & Does the sequence "\{\{substring\}\}" appear in the word "\{\{word\}\}"?                                                                                                                                                                                                                                                                                                                                                                                                                                                                                                                                                                                      \\
\bf Human Prompt III       & Does the word "\{\{word\}\}" contain the sequence of letters "\{\{substring\}\}"?                                                                                                                                                                                                                                                                                                                                                                                                                                                                                                                                                                             \\
\bf Human Prompt IV        & Is "\{\{substring\}\}" present as a substring in the word "\{\{word\}\}"?                                                                                                                                                                                                                                                                                                                                                                                                                                                                                                                                                                                     \\\midrule
\bf Claude Improved Prompt & \begin{tabular}[c]{@{}l@{}}You are tasked with determining whether a given substring is part of a specified word. Here are the inputs:\\ \\ \textless{}substring\textgreater{}\{\{substring\}\}\textless{}/substring\textgreater\\ \textless{}word\textgreater{}\{\{word\}\}\textless{}/word\textgreater\\ \\ Example output structure:\\ {[}Yes/No{]}, the substring {[}is/is not{]} part of the word.\\ \\ Please provide your answer based on the given inputs.\end{tabular}                                                                                                                                                                               \\\midrule\midrule
\multicolumn{2}{c}{\bf Task III: Word Len}                                                                                                                                                                                                                                                                                                                                                                                                                                                                                                                                                                                                                                                 \\
\bf Original Prompt        & How many characters in the word “\{\{word\}\}”?                                                                                                                                                                                                                                                                                                                                                                                                                                                                                                                                                                                                               \\\midrule
\bf Human Prompt I         & What is the total number of characters in the word "\{\{word\}\}"?                                                                                                                                                                                                                                                                                                                                                                                                                                                                                                                                                                                            \\
\bf Human Prompt II        & Can you count the letters in the word "\{\{word\}\}"?                                                                                                                                                                                                                                                                                                                                                                                                                                                                                                                                                                                                         \\
\bf Human Prompt III       & Could you tell me how many letters are in the word "\{\{word\}\}"?                                                                                                                                                                                                                                                                                                                                                                                                                                                                                                                                                                                            \\
\bf Human Prompt IV        & How many alphabetic characters does the word "\{\{word\}\}" contain?                                                                                                                                                                                                                                                                                                                                                                                                                                                                                                                                                                                          \\\midrule
\bf Claude Improved Prompt & \begin{tabular}[c]{@{}l@{}}You are tasked with counting the number of characters in a given word. \\ \\ Here is the word:\\ \textless{}word\textgreater\\ \{\{word\}\}\\ \textless{}/word\textgreater\\ \\ Note: Include all characters in your count, including letters, numbers, and any special characters that may be present.\end{tabular}                                                                                                                                                                                                                                                                                                               \\\midrule\midrule
\multicolumn{2}{c}{\bf Task IV: Distinct Char}                                                                                                                                                                                                                                                                                                                                                                                                                                                                                                                                                                                                                                             \\
\bf Original Prompt        & How many distinct characters in the word “\{\{word\}\}”?                                                                                                                                                                                                                                                                                                                                                                                                                                                                                                                                                                                                      \\\midrule
\bf Human Prompt I         & How many different letters are found in the word "\{\{word\}\}"?                                                                                                                                                                                                                                                                                                                                                                                                                                                                                                                                                                                              \\
\bf Human Prompt II        & What is the number of unique letters present in the word "\{\{word\}\}"?                                                                                                                                                                                                                                                                                                                                                                                                                                                                                                                                                                                      \\
\bf Human Prompt III       & What's the count of distinct characters in the word "\{\{word\}\}"?                                                                                                                                                                                                                                                                                                                                                                                                                                                                                                                                                                                           \\
\bf Human Prompt IV        & Can you identify how many unique characters make up the word "\{\{word\}\}"?                                                                                                                                                                                                                                                                                                                                                                                                                                                                                                                                                                                  \\\midrule
\bf Claude Improved Prompt & \begin{tabular}[c]{@{}l@{}}You are tasked with counting the number of distinct characters in a given word. Here is the word:\\ \\ \textless{}word\textgreater\\ \{\{word\}\}\\ \textless{}/word\textgreater\\ \\ \\ Your final response should consist of only the integer representing the count of distinct characters.\end{tabular}                                                                                                                                                                                                                                                                                                                        \\ \bottomrule
\end{tabular}%
}
\caption{Diverse prompts used to evaluate impact of prompt engineering on LLM performance. Human prompts are brief while prompts improved by Claude contain more detailed instructions and response format requirements.}
\label{tab:prompt_engineering_prompt}
\end{table*}
% \usepackage{graphicx}
\begin{table*}[t!]
\centering
% \resizebox{\textwidth}{!}{%
\begin{tabular}{lcccccc}\toprule
\textbf{Models} & \textbf{Original} & \textbf{Humnan I} & \textbf{Human II} & \textbf{Human III} & \textbf{Human IV} & \textbf{Claude} \\\midrule
\multicolumn{7}{c}{\textbf{Task I: Char Occur}}                                                                                        \\
GPT-4o          & 82.4              & 91.6              & 89.6              & 91.2               & 93.4              & 85.2            \\
Llama-3         & 34.6              & 26.4              & 55.4              & 24.6               & 40.4              & 22.4            \\\midrule
\multicolumn{7}{c}{\textbf{Task II: Substring Occur}}                                                                                  \\
GPT-4o          & 87.4              & 94.7              & 89                & 94                 & 87.4              & 88.1            \\
Llama-3         & 58.2              & 71.7              & 59.7              & 65.3               & 61.4              & 52.8            \\\midrule
\multicolumn{7}{c}{\textbf{Task III: Word Len}}                                                                                        \\
GPT-4o          & 92.0              & 92.4              & 96.2              & 95.4               & 96.8              & 97.6            \\
Llama-3         & 74.6              & 92.6              & 91.0              & 79.2               & 77.8              & 96.8            \\\midrule
\multicolumn{7}{c}{\textbf{Task IV: Distinct Char}}                                                                                    \\
GPT-4o          & 89.2              & 87.2              & 78.0              & 90.6               & 95.8              & 52.8            \\
Llama-3         & 57.8              & 56.8              & 82.6              & 63.2               & 43.2              & 21.6 \\\bottomrule          
\end{tabular}%
% }
\caption{Impact of prompt engineering on word-based counting tasks. Neither human heuristics-driven prompts nor improved prompts from Claude are able to consistently elicit better capabilities from LLMs for solving word-based counting problems.}
\label{tab:prompt_engineering_results}
\end{table*}
\subsection{Related Work}\label{sec:related_work}
\paragraph{Failure Modes of LLMs}
Although LLMs have exhibited strong capabilities to complete tasks requiring extensive world knowledge and complex reasoning, they still present some unexpected failures. \citet{berglund2023reversal} discovered the \emph{reversed curve}, where an LLM that recognizes ``A is B'' does not necessarily learn that ``B is A.'' Another challenging posed to LLMs is irrelevant context, which distracts models from completing tasks as normal. For instance, \citet{shi2023large} found that adding irrelevant context in the problem statement leads to a noticeable performance drop on multiple reasoning benchmarks. Moreover, \citet{chen2024premise} show that including irrelevant rules degrades the logical reasoning performance of LLMs. Sensitivity to text order is another challenge that LLMs struggle with. For example, \citet{chen2024premise} observed that in deductive reasoning tasks, presenting the premises in the same order as the ground truth proof in the prompt (as opposed to random ordering) drastically increases the model’s accuracy, while permuting the premise order can cause a performance drop of over $30\%$. 
% Extending context lengths during pretraining has enabled LLMs to solve ``needle in haystack''~\citep{LLMTest_NeedleInAHaystack}, where the goal is to find a specific string in a long text. 
Another example is the lost-in-the-middle phenomenon in the long-context scenario, in which LLM performance drops drastically when they need to utilize input context in the middle rather than that in the beginning or the end~\citep{liu-etal-2024-lost}.

% Compared with prior challenges that LLMs are not capable to handle, word-based counting is far simpler, only requiring highly elementary understanding of characters and simple counting.
\paragraph{Word-based Counting}
Failure to count the number of specific character within the queried word is a recently emergent problem that most LLMs struggle with~\citep{karpathy_tweet_2024_tokenization}. \citet{yehudai2024can} attributed such deficiency to constraints from LLM architecture, emphasizing that it is likely impossible for a size-limited transformer to complete the counting task. \citet{ball2024can} examined capabilities of GPT-4 on character occurrence task and showed sensitivity of task-accuracy both to query phrasing and input parameter population. \citet{shin2024large} observed significant performance contrast between character and token (i.e. subword) input. They also proposed the tokenization issue and lack of training on similar data as potential reasons for such failure.

Different from prior literature that focuses on demonstrating the failure mode or proposing potential reasons, we carefully design multiple evaluation settings and empirically show invalidness of major conjectures. More importantly, we investigate promising strategies and show that reasoning is a promising direction to solve word-based counting problems.  
\begin{table}[t!]
\centering
% \resizebox{\linewidth}{!}{%
\begin{tabular}{@{}lllll@{}}
\toprule
Task                                                          & Attribute                                                                  & Min & Max & Avg. \\ \midrule
\begin{tabular}[c]{@{}l@{}}\textbf{I: Char} \\ \textbf{Occur}\end{tabular}      & \begin{tabular}[c]{@{}l@{}}Occurence \\ of asked \\ character\end{tabular} & 1   & 4   & 1.22 \\\midrule
\begin{tabular}[c]{@{}l@{}}II: \textbf{Substring}\\ \textbf{Occur}\end{tabular} & \begin{tabular}[c]{@{}l@{}}Length of \\ substring\end{tabular}             & 3   & 14  & 5.39 \\\midrule
\textbf{III: Word Len}                                                 & \begin{tabular}[c]{@{}l@{}}Number of\\ characters\end{tabular}             & 3   & 18  & 9.34 \\\midrule
\begin{tabular}[c]{@{}l@{}}\textbf{IV: Distinct} \\ \textbf{Char}\end{tabular}  & \begin{tabular}[c]{@{}l@{}}Number of\\ distinct\\ characters\end{tabular}  & 3   & 13  & 7.50 \\ \bottomrule
\end{tabular}%
% }
\caption{Statistics of evaluated tasks. In each row, we list information of key component to each task. We randomly sample $500$ instances for Task I, III and IV, while preparing a balanced dataset with $500$ positive and $500$ negative instances for Task II.}
\label{tab:dataset_statistics}
\end{table}

\subsection{Experimental Setup}~\label{sec:experimental_setup_appendix}
\paragraph{Language Models}
For comprehensive evaluation of LLMs capabilities on simple word-based counting problems, we consider $9$ prevalent families of powerful instructed or chat models including both open-source and proprietary ones: Llama 3 (8B-instruct)~\citep{dubey2024llama}, Qwen 1.5 (7B-chat)~\citep{bai2023qwen}, Gemma 1 (7B-instruct)~\citep{team2024gemma}, InternLM2 (7B-chat)~\citep{cai2024internlm2}, Phi 3 (small-128k-instruct)~\citep{abdin2024phi}, Mistral v0.3 (7B-instruct)~\citep{jiang2023mistral}, DeepSeek V2 (Lite-chat)~\citep{deepseekv2}, Yi 1.5 (9B-chat)~\citep{young2024yi}, and GPT-4o~\citep{openai_gpt4o}. Unless otherwise stated, we follow prior benchmark literature~\citep{suzgun2022challenging,zhong-etal-2024-agieval} by adopting greedy decoding~\footnote{We set the maximum number of tokens for generation to 128.} to minimize the noise for open-ended text generation. We list the checkpoint resource of tested open-source LLMs in~\Cref{tab:llm_info}. 
\paragraph{Evaluation Metrics}
For Task II Substring Cccur where the ground-truth answer is ``Yes'' or ``No'', we measure accuracy using \emph{soft match}, computed by checking whether the true answer appears in models' responses or not. For the other three tasks where models are expected to answer a number, we extract the last digits from model responses automatically and examine whether they are identical to the true answer. We also consider the verbal representation of numbers (e.g., ``two'' and ``twice'' for ``2'') with soft match by comparing the generated output with word form of true numbers.
% 1) Qwen2 Math~\citep{qwen2_math} and CodeQwen 1.5~\citep{codeqwen1.5}, 2) CodeGemma~\citep{team2024codegemma}, 3) InternLM2 Math Plus~\citep{ying2024internlm} and InternLM2 Step Prover~\citep{wu2024leangithubcompilinggithublean}, 4) Mathstral v0.1~\citep{mistral_mathstral} and Codestral v0.1~\citep{mistral_codestral}, 5) DeepSeekMath~\citep{shao2024deepseekmath}, DeepSeek Prover V1.5~\citep{xin2024deepseek} and DeepSeek Coder V2~\citep{zhu2024deepseek}, 6) Yi Coder~\citep{01ai_llm_for_code}.
\section{Prompt Engineering}\label{sec:prompt_engineering}

In~\Cref{tab:prompt_engineering_prompt}, we list human prompts and improved prompts from Claude~\footnote{ Prompt improver from Claude introduced the ability to improve prompts for better completions, which takes existing prompts and leverages Claude to automatically refine them using advanced prompt engineering techniques. NOTE that we have removed chain-of-thought reasoning instruction and examples in the improved prompts to form fair comparison with original simple prompts.} on four counting tasks. Different from concise human prompts, prompts automatically improved by Claude normally contain detailed instructions and explicitly specify expected formats of responses.

We present LLM performance in response to different task prompts in~\Cref{tab:prompt_engineering_results}. In contrast to stable performance enhancement  from reasoning-based strategies, neither human heuristics-driven  nor automatically improved prompts from Claude are able to elicit better capabilities to address the studied word-based counting problems. We conclude that failures of LLMs for solving counting tasks are irrelevant to how we phrase prompts.
\section{Whether Math or Code Training Data Helps}
\subsection{Setup}\label{sec:whether_setup_more}
\paragraph{Math/Code Models} We compare LLMs fine-tuned on general instruction/chat data (described in~\Cref{sec:exp_setup}) with their counterparts specialized in math- or code-related tasks: 1) Qwen2 Math~\citep{qwen2_math} and CodeQwen 1.5~\citep{codeqwen1.5}, 2) CodeGemma~\citep{team2024codegemma}, 3) InternLM2 Math Plus~\citep{ying2024internlm} and InternLM2 Step Prover~\citep{wu2024leangithubcompilinggithublean}, 4) Mathstral v0.1~\citep{mistral_mathstral} and Codestral v0.1~\citep{mistral_codestral}, 5) DeepSeekMath~\citep{shao2024deepseekmath}, DeepSeek Prover V1.5~\citep{xin2024deepseek} and DeepSeek Coder V2~\citep{zhu2024deepseek}, 6) Yi Coder~\citep{01ai_llm_for_code}. We list detailed model information in~\Cref{tab:llm_info}.
\paragraph{Implementations}
Besides prompting LLMs to answer the word-based counting tasks defined in~\Cref{sec:exp_setup} in the open-ended setting, we also explicitly request code LLMs to generate Python codes~\footnote{We adopt few-shot prompting used in~\citep{gao2023pal,guo2024deepseek} so that models follow code formats demonstrated in provided examples, which makes code extraction easier and evaluation more accurate.}. We then measure correctness by executing codes and comparing output with the ground-truth.
\section{How to Make LLMs Experts Again}
\subsection{Implementations}~\label{sec:expert_implementation}
We use greedy decoding and adopt the zero-shot setting~\footnote{To avoid impact of demonstrations on generation, we use zero-shot rather than few-shot to evaluate effectiveness of different reasoning strategies, where the expected format of responses is included in questions.} for model generation as introduced in~\Cref{sec:exp_setup} for most strategies. For \emph{self-consistency} and \emph{ToT}, we follow the practice in literature~\citep{wang2022self,yao2024tree} by applying temperature sampling with $T=0.7$ and truncating at the top-k ($k=40$) tokens with the highest probability, we set the reasoning path to $5$ unless otherwise specified. We finetune Llama 3 with Lora~\citep{hu2021lora} on $10,000$ training instances and set the learning rate to $3e-4$, epoch to $1$ and batch size to $128$ on a single A100 80G device~\footnote{Considering limited GPU memory, we employ the batch size of $2$ and set the gradient accumulation steps to $64$ in practice.)}. We also measure the impact of finetuning on existing capabilities with finetuned models evaluated on general (MMLU and IFEval), reasoning (GPQA and Hellaswag), math (GSM8K)  and coding (HumanEval) benchmarks following~\citep{llama2024}. For ICL, we randomly sample 4 and 8 demonstrations per testing instance from the training set.
\begin{table*}[t!]
\centering
\resizebox{2\columnwidth}{!}{%
\begin{tabular}{@{}lcc@{}}
\toprule
LLMs        &\#Params & Download Links/Version                                                    \\ \midrule
Llama 3   & 8B        & \url{https://huggingface.co/meta-llama/Meta-Llama-3-8B-Instruct}             \\\midrule
Qwen 1.5 & 7B       & \url{https://huggingface.co/Qwen/Qwen1.5-7B-Chat}            \\
Qwen2 Math&7B&\url{https://huggingface.co/Qwen/Qwen2-Math-7B-Instruct}\\
CodeQwen 1.5&7B&\url{https://huggingface.co/Qwen/CodeQwen1.5-7B-Chat}\\\midrule

Gemma 1       & 7B         & \url{https://huggingface.co/google/gemma-7b-it}                      \\
CodeGemma&7B&\url{https://huggingface.co/google/codegemma-7b-it}\\\midrule
InternLM2      & 7B        & \url{https://huggingface.co/internlm/internlm2-chat-7b}                     \\
InternLM2 Math Plus&7B&\url{https://huggingface.co/internlm/internlm2-math-plus-7b}\\
InternLM2 Step Prover&7B&\url{https://huggingface.co/internlm/internlm2-step-prover}\\\midrule
Phi 3     &     7B    & \url{https://huggingface.co/microsoft/Phi-3-small-128k-instruct} \\
Mistral v0.3    & 7B        & \url{https://huggingface.co/mistralai/Mistral-7B-Instruct-v0.3}                \\
Mathstral v0.1&7B&\url{https://huggingface.co/mistralai/Mathstral-7B-v0.1}\\
Codestral v0.1&22B&\url{https://huggingface.co/mistralai/Codestral-22B-v0.1}\\\midrule
DeepSeek-V2      &   16B     & \url{https://huggingface.co/deepseek-ai/DeepSeek-V2-Lite-Chat}    \\
DeepSeekMath&7B&\url{https://huggingface.co/deepseek-ai/deepseek-math-7b-rl}\\
DeepSeek Prover V1.5&7B&\url{https://huggingface.co/deepseek-ai/DeepSeek-Prover-V1.5-RL}\\
DeepSeek Coder V2&16B&\url{https://huggingface.co/deepseek-ai/DeepSeek-Coder-V2-Lite-Instruct}\\\midrule
Yi 1.5     & 9B        & \url{https://huggingface.co/01-ai/Yi-1.5-9B-Chat}   \\
Yi Coder&9B&\url{https://huggingface.co/01-ai/Yi-Coder-9B-Chat}\\\midrule
GPT-4o &    -   & gpt-4o-2024-05-13                  \\ \bottomrule
\end{tabular}%
}
% \vspace{-0.5em}
\caption{Information of tested LLMs. We list their model sizes and the download links if available  or the model version for the proprietary model.}
% \vspace{-1em}
\label{tab:llm_info}
\end{table*}

\begin{table*}[t!]
\centering
\resizebox{\textwidth}{!}{%
\begin{tabular}{@{}lccccccccc@{}}
\toprule
\textbf{Task}       & \textbf{GPT-4o} & \textbf{Llama 3} & \textbf{Qwen 1.5} & \textbf{Gemma 1} & \textbf{InternLM2} & \textbf{Phi 3} & \textbf{Mistral v0.3} & \textbf{DeepSeek V2} & \textbf{Yi 1.5} \\ \midrule
\multicolumn{10}{c}{\textbf{Germanic Languages}}                                                                                                                                                       \\
\textbf{English}    & 82.4            & 34.6             & 30.6              & 41.2             & 60.8               & 39.0           & 35.4                  & 27.2                 & 46.6            \\
\textbf{German}     & 69.6            & 27.2             & 20.6              & 5.2              & 50.6               & 40.2           & 21.8                  & 34.8                 & 38.4            \\
\textbf{Swedish}    & 80.6            & 39.0             & 18.8              & 5.8              & 61.2               & 38.0           & 34.4                  & 55.6                 & 42.6            \\\midrule
\multicolumn{10}{c}{\textbf{Romance Languages}}                                                                                                                                                        \\
\textbf{French}     & 75.6            & 38.0             & 16.4              & 10.0             & 63.6               & 45.0           & 26.0                  & 40.2                 & 52.6            \\ 
\textbf{Spanish}    & 76.2            & 32.6             & 25.4              & 10.4             & 64.6               & 45.6           & 28.0                  & 38.2                 & 50.4            \\
\textbf{Italian}    & 71.4            & 24.8             & 22.0              & 15.6             & 55.2               & 37.6           & 20.4                  & 37.0                 & 49.6            \\
\textbf{Portuguese} & 65.0            & 31.2             & 21.4              & 23.0             & 65.4               & 52.6           & 26.0                  & 45.8                 & 47.4  \\\bottomrule         
\end{tabular}%
}
\caption{Performance of LLMs on \emph{Task I Char Occur} in different languages from Germanic and Romance language families. LLMs cannot better identify occurrence of characters in less common words.}
\label{tab:lan_performance}
\end{table*}

\begin{table*}[t!]
\centering
% \resizebox{\textwidth}{!}{%
\begin{tabular}{@{}lcl@{}}
\toprule
\textbf{Perturbation}                                            & \textbf{Perturbed Word}                                         & \textbf{Description}                                                 \\ \midrule
delete                                                           & straberry                                                       & "w" is deleted                                                       \\
insert                                                           & strawbe\textcolor{red}{k}rry                                                     & "k" is inserted                                                      \\
repeat                                                           & st\textcolor{red}{t}rawberry                                                     & "t" is repeated                                                      \\
replace                                                          & str\textcolor{red}{s}wberry                                                      & "a" replaced with "s"                                                \\
swap                                                             & str\textcolor{red}{y}wberr\textcolor{red}{a}                                                      & "a" and ''y" are swapped                                             \\
left shift                                                       & trawberrys                                                      & all letters shift left with the first letter "s" moving to the end   \\
right shift                                                      & ystrawberr                                                      & all letters shift right with the last letter "y" moving to the start \\
shuffle                                                          & rasbretyrw                                                      & all letters arranged in random order                                 \\
\begin{tabular}[c]{@{}l@{}}mapping\\ (alphabetical)\end{tabular} & abcdefghhi & letters from left to right replaced by "a", "b", "c", etc.           \\
\begin{tabular}[c]{@{}l@{}}mapping \\ (special)\end{tabular}     & !@\#\$\%\&'(()                                                  & letters from left to right replaced by "!", "@", "\#", etc.          \\
+dash                                                            & s\textcolor{red}{-}t\textcolor{red}{-}r\textcolor{red}{-}a\textcolor{red}{-}w\textcolor{red}{-}b\textcolor{red}{-}e\textcolor{red}{-}r\textcolor{red}{-}r\textcolor{red}{-}y                                             & dash '-' inserted between every two letters                          \\
+space                                                           & s t r a w b e r r y                                             & space ' ' inserted between every two letters                         \\
+comma                                                           & s\textcolor{red}{,}t\textcolor{red}{,}r\textcolor{red}{,}a\textcolor{red}{,}w\textcolor{red}{,}b\textcolor{red}{,}e\textcolor{red}{,}r\textcolor{red}{,}r\textcolor{red}{,}y                                             & comma ',' inserted between every two letters                         \\ \bottomrule
\end{tabular}%
% }
\caption{Character-level perturbation examples on the word "strawberry" when the question is "How many r's in the word "strawberry"?"}
\label{tab:implicit_tokenization}
\end{table*}

\begin{figure*}[t!]
     \centering
         \begin{subfigure}[b]{\textwidth}
         \centering
         \includegraphics[width=\linewidth]{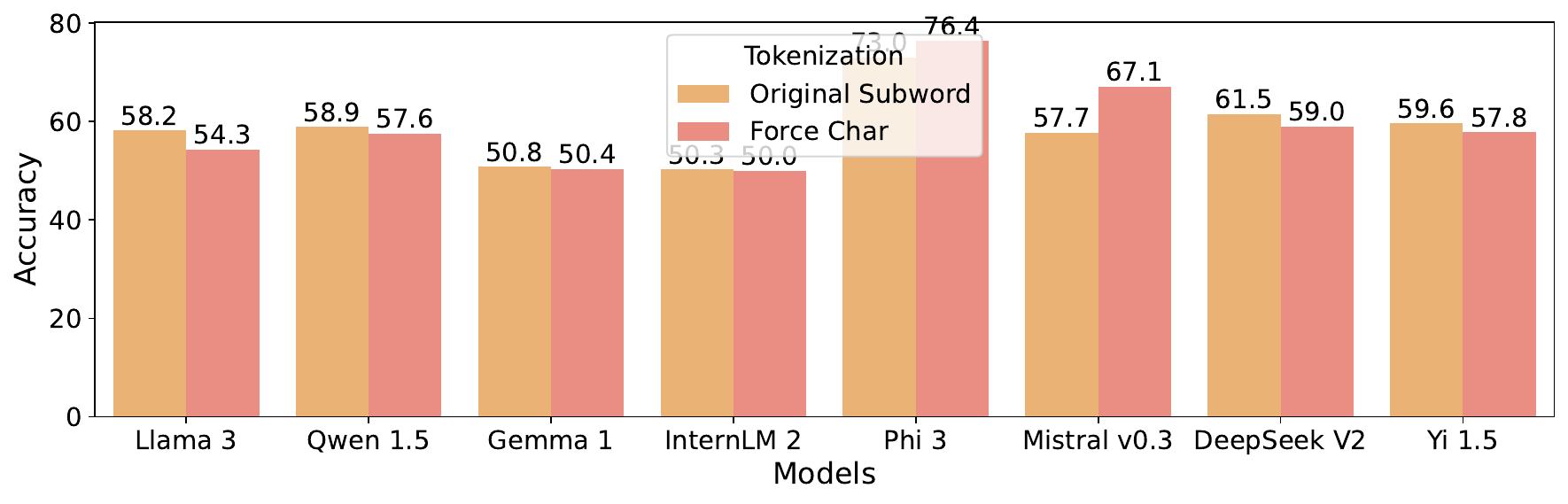}\caption{Task II Substring Occur.}
         \label{fig:force_tokenization_strawberry_substring_occurrence}
     \end{subfigure}
              \begin{subfigure}[b]{\textwidth}
         \centering
         \includegraphics[width=\linewidth]{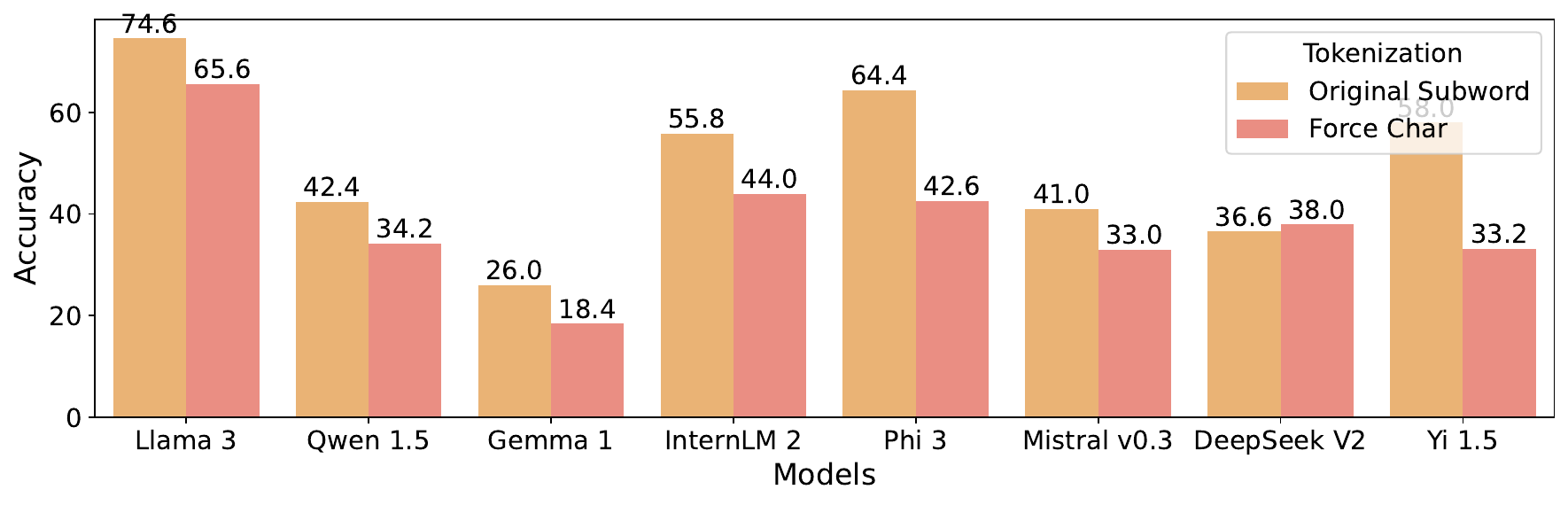}\caption{Task III Word Len.}
         \label{fig:force_tokenization_strawberry_word_length}
     \end{subfigure}
              \begin{subfigure}[b]{\textwidth}
         \centering
         \includegraphics[width=\linewidth]{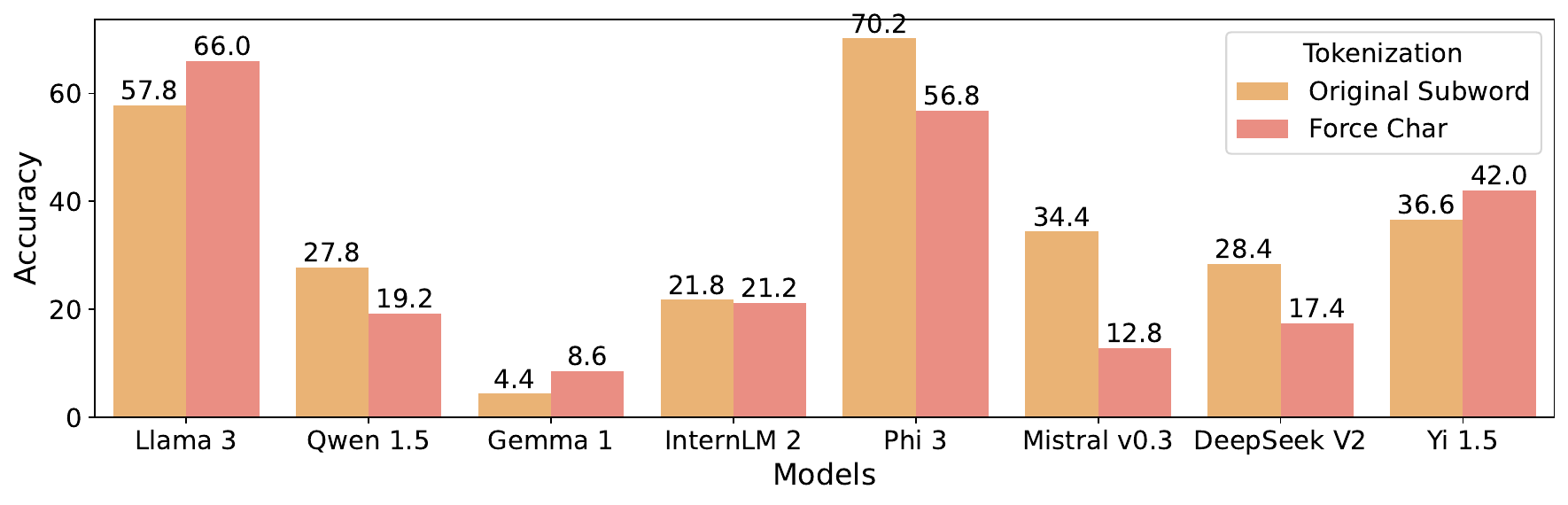}\caption{Task IV Distinct Char.}
         \label{fig:force_tokenization_strawberry_distinct_characters}
     \end{subfigure}
        \caption{Performance comparison on three other word-based counting tasks to further verify that \textbf{Conjecture I} is incorrect.}
        \label{fig:tokenization_extra}
        % \vspace{-1em}
\end{figure*}
\begin{figure*}[t!]
     \centering
    \begin{subfigure}[b]{\textwidth}
         \centering
         \includegraphics[width=\linewidth]{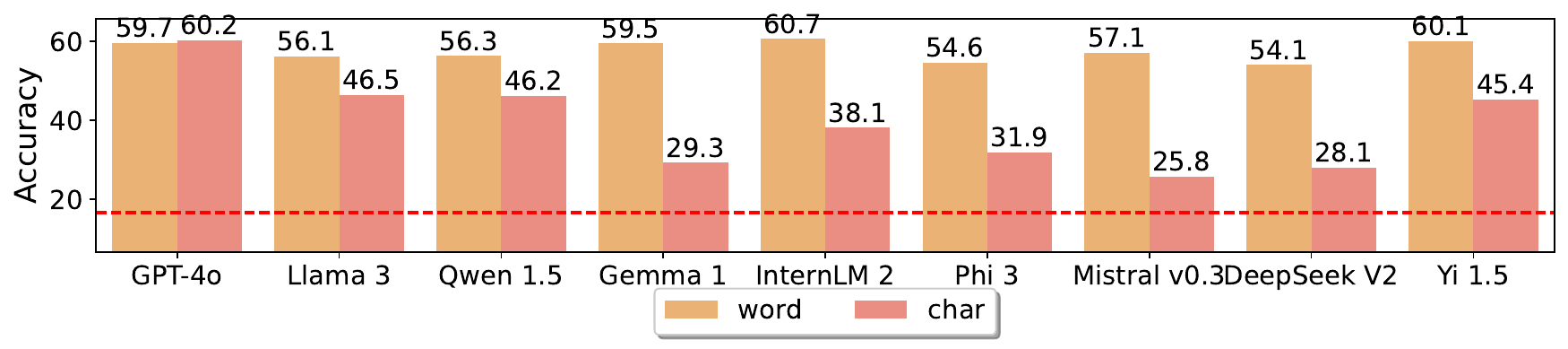}\caption{Emotion.}
         \label{fig:emotion}
     \end{subfigure}
     %     \begin{subfigure}[b]{\textwidth}
     %     \centering
     %     \includegraphics[width=\linewidth]{classification_imdb.pdf}\caption{IMDB.}
     %     \label{fig:imdb}
     % \end{subfigure}
         \begin{subfigure}[b]{\textwidth}
         \centering
         \includegraphics[width=\linewidth]{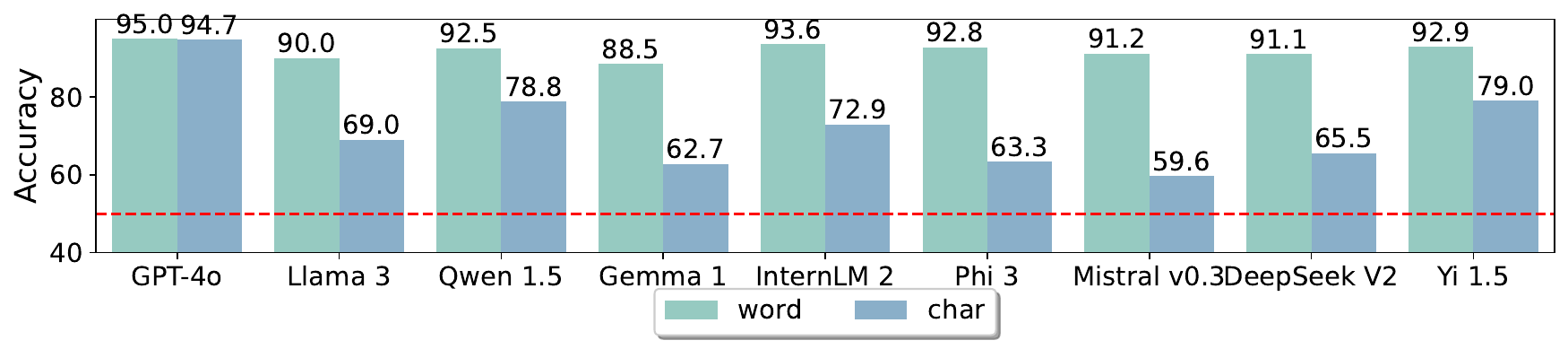}\caption{SST-2.}
         \label{fig:sst2}
     \end{subfigure}
        \caption{Performance comparison of LLMs between natural word and character input on two classification datasets.}
        \label{fig:classification_more}
        % \vspace{-1em}
\end{figure*}
\begin{figure*}[t!]
     \centering
     % \includegraphics[width=\linewidth]{length_unique_gpt-4o.pdf}%\caption{GPT-4o.}
    % \begin{subfigure}[b]{\textwidth}
    %      \centering
    %      \includegraphics[width=\linewidth]{length_unique_gpt-4o.pdf}%\caption{GPT-4o.}
    %      \label{fig:unique_len_gpt_4o}
    %  \end{subfigure}
    %      \begin{subfigure}[b]{\textwidth}
    %      \centering
         \includegraphics[width=\linewidth]{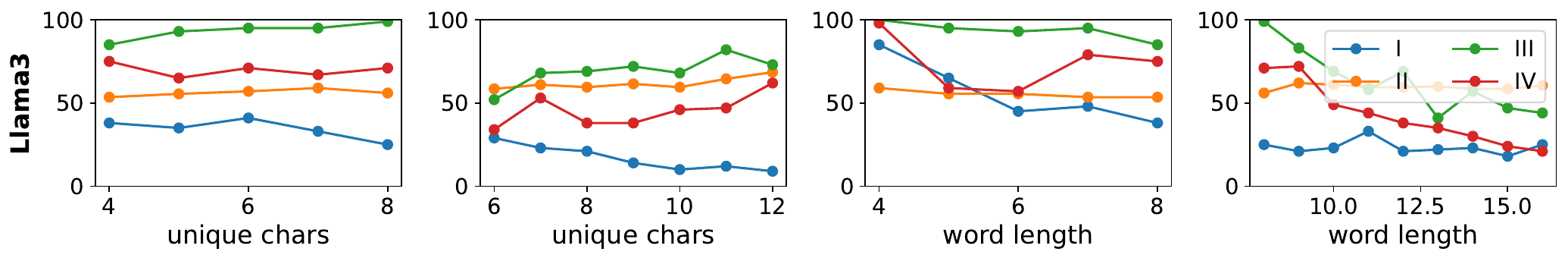}%\caption{Llama 3.}
    %      \label{fig:unique_len_llama3}
    %  \end{subfigure}
        \caption{Performance variation from Llama 3 on four tasks.}
        \label{fig:unique_len_more}
        % \vspace{-1em}
\end{figure*}

\begin{figure*}[t!]
     \centering
         \begin{subfigure}[b]{\textwidth}
         \centering
         \includegraphics[width=\linewidth]{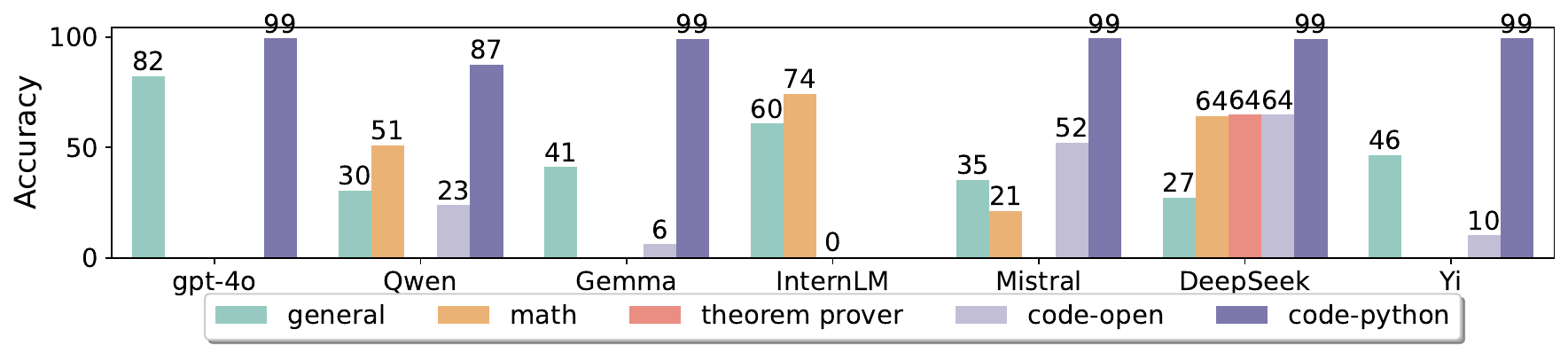}\caption{Task I: Char Occur.}
         \label{fig:math_code_strawberry_text}
     \end{subfigure}
    \begin{subfigure}[b]{\textwidth}
         \centering
         \includegraphics[width=\linewidth]{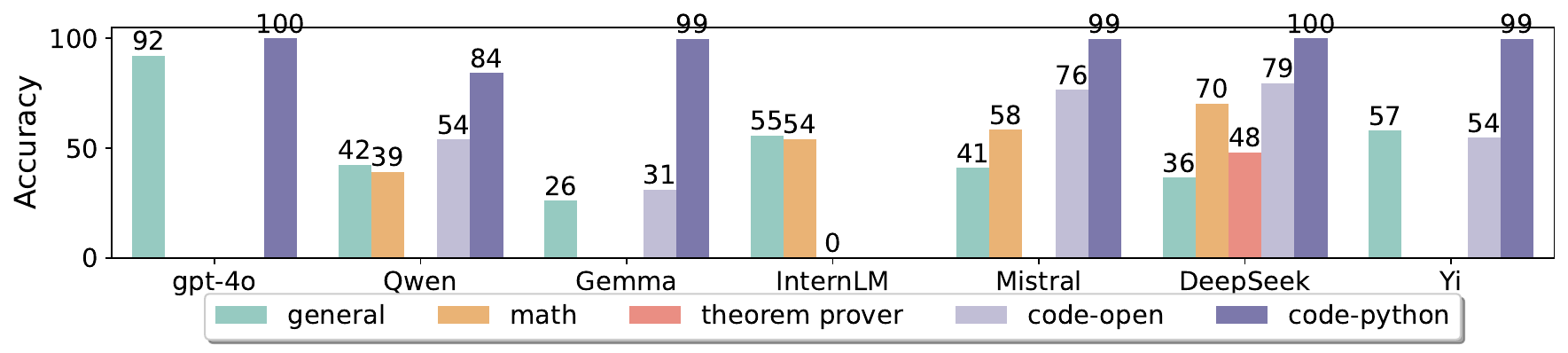}\caption{Task III: Word Len.}
         \label{fig:math_code_strawberry_word_length}
     \end{subfigure}
         \begin{subfigure}[b]{\textwidth}
         \centering
         \includegraphics[width=\linewidth]{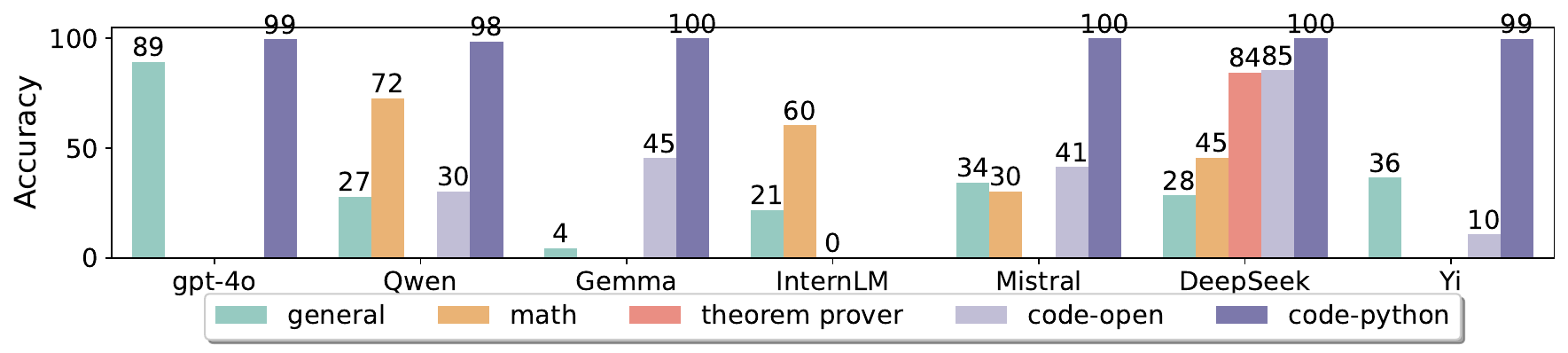}\caption{Task IV: Distinct Char.}
         \label{fig:math_code_strawberry_distinct_characters}
     \end{subfigure}
        \caption{Performance of math and code LLMs on three word-based counting tasks.}
        \label{fig:math_code_more}
        % \vspace{-1em}
\end{figure*}
\begin{figure*}[t!]
     \centering
    \begin{subfigure}[b]{\textwidth}
         \centering
         \includegraphics[width=\linewidth]{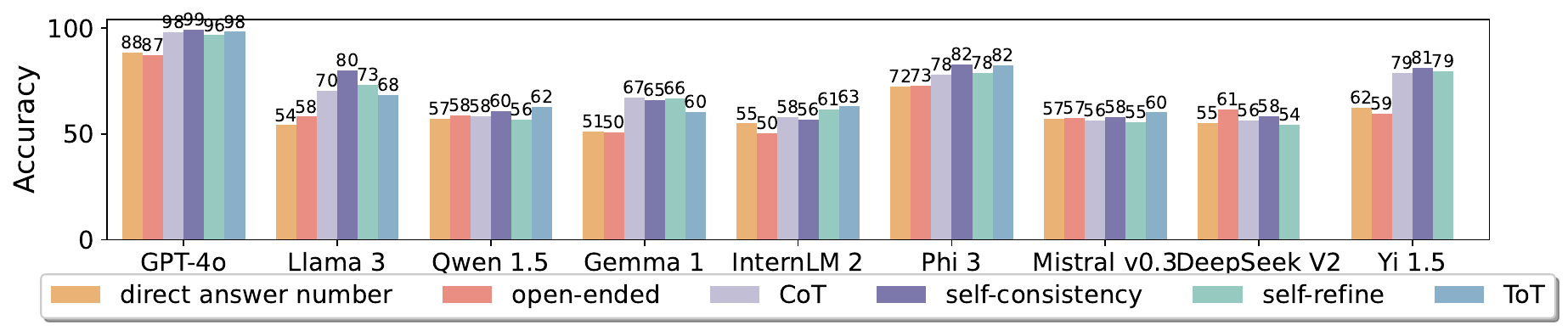}\caption{Task II: Substring Occur.}
         \label{fig:reasoning_substring_occur}
     \end{subfigure}
         \begin{subfigure}[b]{\textwidth}
         \centering
         \includegraphics[width=\linewidth]{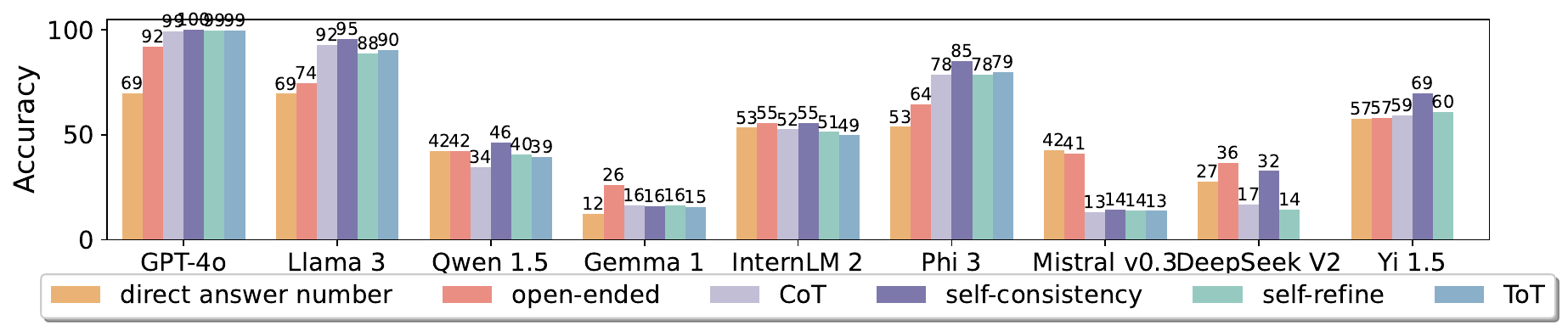}\caption{Task III: Word Len.}
         \label{fig:reasoning_word_len}
     \end{subfigure}
         \begin{subfigure}[b]{\textwidth}
         \centering
         \includegraphics[width=\linewidth]{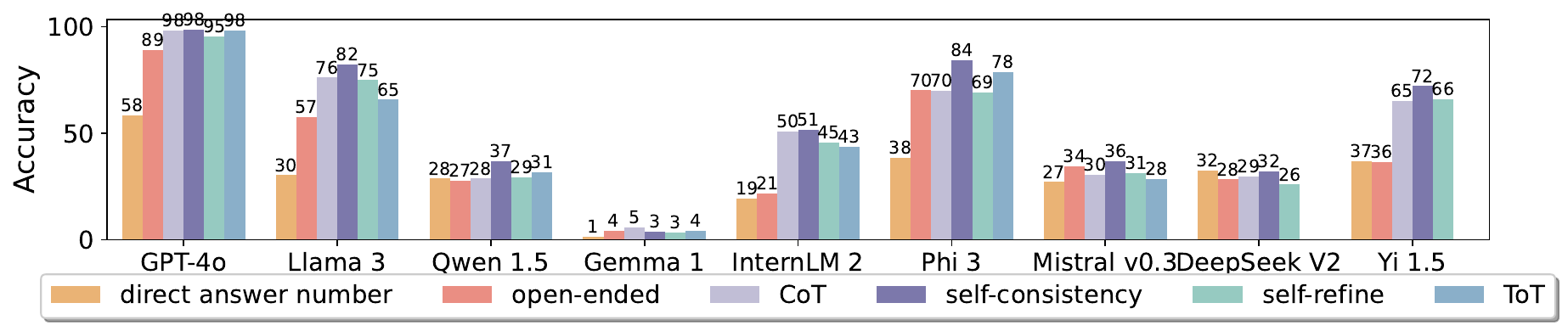}\caption{Task IV: Distinct Char.}
         \label{fig:reasoning_distinct_char}
     \end{subfigure}
        \caption{Impact of different reasoning strategies on LLM performance for three word-based counting tasks.}
        \label{fig:reasoning_others}
        % \vspace{-1em}
\end{figure*}
\begin{figure*}[t!]
     \centering
     \includegraphics[width=\linewidth]{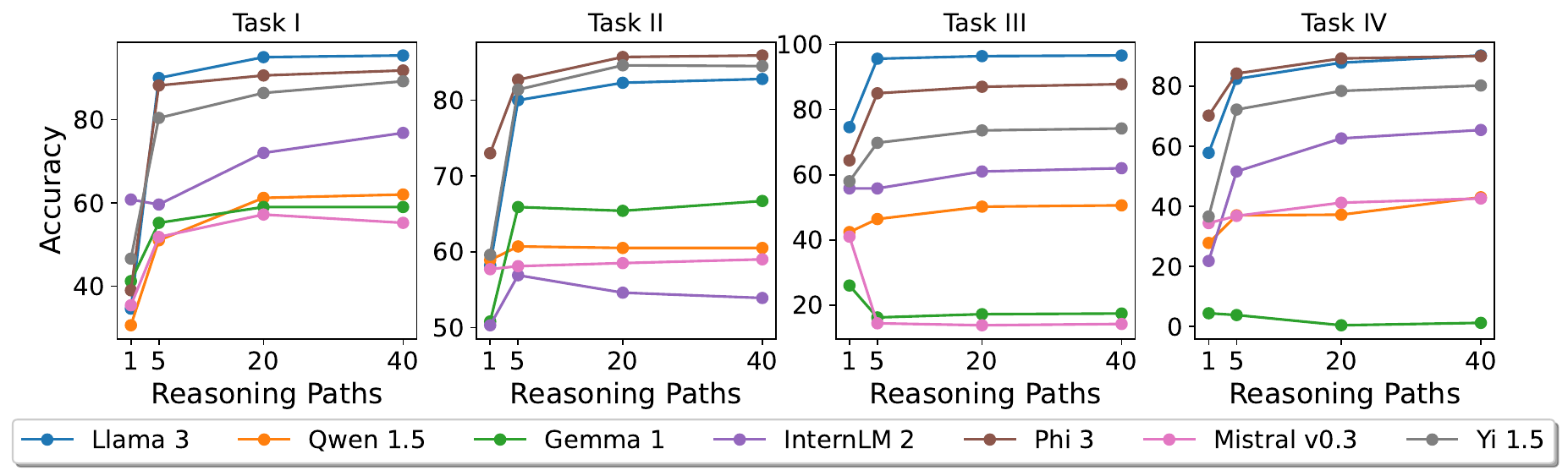}
        \caption{Scaling trend of the \emph{self-consistency} strategy by different LLMs on four tasks.}
        \label{fig:scaling}
        % \vspace{-1em}
\end{figure*}

\begin{table}[t!]
\centering
\resizebox{\linewidth}{!}{%
\begin{tabular}{@{}l@{}}
\toprule
% Text I: Char Occur                                                                                                                                                                                                                                                                                                                                                                                                                                           \\ \midrule
\textbf{Question:} How many a's in the word "confusable"?                                                                                                                                                                                                                                                                                                                                                                                                          \\\midrule
\textbf{True Answer:} 1                                                                                                                                                                                                                                                                                                                                                                                                                                               \\\midrule
\textbf{Direct Answer Number:} 2 \textbf{\textcolor{red}{\ding{56}
}}\\\midrule
\textbf{Open-ended:} There are 2 a's in the word "confusable". \textbf{\textcolor{red}{\ding{56}
}}                                                                                                                                                                                                                                                                                                                                                                                             \\\midrule
\begin{tabular}[c]{@{}l@{}}\textbf{CoT:} 
Here's the step-by-step reasoning:\\\\

Reasoning:\\
1. Start by looking at the word "confusable".\\
2. Identify each letter in the word: c-o-n-f-u-s-a-b-l-e.\\
3. Count the number of "a"s in the word: there is one\\ "a" in the word.\\\\

Answer:\\
1 \textbf{\textcolor{blue}{\ding{52}}}
\end{tabular} \\ \bottomrule
\end{tabular}%
}
\caption{Case study of baseline and CoT generation from Llama 3 on \emph{Task I Char Occur}. We mark correct answer by \textbf{\textcolor{blue}{\ding{52}}} and wrong ones by \textbf{\textcolor{red}{\ding{56}}}. We provide examples on other three datasets in~\Cref{tab:case_study_others}.}
\label{tab:case_study_strawberry_text}
        % \vspace{-1em}
\end{table}

\begin{table*}[t!]
\centering
\resizebox{\textwidth}{!}{%
\begin{tabular}{@{}l@{}}
\toprule
\multicolumn{1}{c}{\bf Task II: Substring Occur}                                                                                                                                                                                                                                                                                                                                                                                                                                                                                                                                                                                                                                                                                                                                                                     \\
\textbf{Question:} Is substring "Elvp" part of the word "Elvis"?                                                                                                                                                                                                                                                                                                                                                                                                                                                                                                                                                                                                                                                                                                                                    \\
\textbf{True Answer:} No.                                                                                                                                                                                                                                                                                                                                                                                                                                                                                                                                                                                                                                                                                                                                                                           \\
\textbf{Direct Answer Yes/No:} Yes. (\textbf{\textcolor{red}{Wrong}})\\
\begin{tabular}[c]{@{}l@{}}\textbf{Open-ended:} A simple question!\\ \\ \\ Yes, the substring "Elvp" is indeed part of the word "Elvis". (\textbf{\textcolor{red}{Wrong}})\end{tabular}                                                                                                                                                                                                                                                                                                                                                                                                                                                                                                                                                                                                                                               \\
\begin{tabular}[c]{@{}l@{}}\textbf{CoT:} Here's the step-by-step reasoning:\\ \\ Reasoning:\\ 1. The word "Elvis" is a sequence of characters: E-L-V-I-S.\\ 2. The substring "Elvp" is a sequence of characters: E-L-V-P.\\ 3. To determine if "Elvp" is a part of "Elvis", we need to check if all the characters in "Elvp" are present in the same order in "Elvis".\\ 4. Comparing the characters, we can see that the first three characters "Elv" are present in both "Elvp" and "Elvis".\\ 5. However, the fourth character "P" is not present in "Elvis", which has an "I" instead.\\ 6. Since the characters in "Elvp" are not present in the same order in "Elvis", "Elvp" is not a part of "Elvis".\\ \\ Answer:\\ No, the substring "Elvp" is not part of the word "Elvis". (\textbf{\textcolor{blue}{Correct}})\end{tabular} \\\midrule
\multicolumn{1}{c}{\bf Task III: Word Len}                                                                                                                                                                                                                                                                                                                                                                                                                                                                                                                                                                                                                                                                                                                                                                         \\ 
\textbf{Question:} How many characters in the word "Missourianism"?                                                                                                                                                                                                                                                                                                                                                                                                                                                                                                                                                                                                                                                                                                                                 \\
\textbf{True Answer:} 13                                                                                                                                                                                                                                                                                                                                                                                                                                                                                                                                                                                                                                                                                                                                                                            \\
\textbf{Direct Answer Number:} 12 (\textbf{\textcolor{red}{Wrong}})\\
\begin{tabular}[c]{@{}l@{}}\textbf{Open-ended:} Let me count them for you!\\ \\ The word "Missourianism" has 12 characters: M-I-S-S-O-U-R-I-A-N-I-S-M. (\textbf{\textcolor{red}{Wrong}})\end{tabular}                                                                                                                                                                                                                                                                                                                                                                                                                                                                                                                                                                                                                                 \\
\begin{tabular}[c]{@{}l@{}}\textbf{CoT:} Here's the step-by-step reasoning:\\ \\ Reasoning:\\ To count the number of characters in the word "Missourianism", I will break it down into individual letters and count them one by one.\\ \\ 1. M\\ 2. i\\ 3. s\\ 4. s\\ 5. o\\ 6. u\\ 7. r\\ 8. i\\ 9. a\\ 10. n\\ 11. i\\ 12. s\\ 13. m\\ \\ Answer:\\ There are 13 characters in the word "Missourianism". (\textbf{\textcolor{blue}{Correct}})\end{tabular}                                                                                                                                                                                                                                                                                                                                                                             \\\midrule
\multicolumn{1}{c}{\bf Task IV: Distinct Char}                                                                                                                                                                                                                                                                                                                                                                                                                                                                                                                                                                                                                                                                                                                                                                     \\
\textbf{Question:} How many distinct characters in the word "mutch"?                                                                                                                                                                                                                                                                                                                                                                                                                                                                                                                                                                                                                                                                                                                               \\
\textbf{True Answer:} 5                                                                                                                                                                                                                                                                                                                                                                                                                                                                                                                                                                                                                                                                                                                                                                             \\
\textbf{Direct Answer Number:} 5 (\textbf{\textcolor{blue}{Correct}})\\
\textbf{Open-ended:} The word "mutch" contains 5 characters: m-u-t-c-h. There are 4 distinct characters: m, u, t, and c. (\textbf{\textcolor{red}{Wrong}})                                                                                                                                                                                                                                                                                                                                                                                                                                                                                                                                                                                                                                        \\
\begin{tabular}[c]{@{}l@{}}\textbf{CoT:} Reasoning:To find the number of distinct characters in the word "mutch", I will start by writing down the word and identifying\\ each character:\\ \\ m-u-t-c-h\\ \\ Next, I will look for any duplicate characters. In this case, I see that there is no repeated character.\\ \\ Finally, I will count the number of unique characters, which are:\\ \\ m, u, t, c, h\\ \\ Answer:\\ 5 (\textbf{\textcolor{blue}{Correct}})\end{tabular}                                                                                                                                                                                                                                                                                                                                                       
\\\bottomrule                                                                                                                                                                                                                                                                                                          
\end{tabular}%
}
\caption{Case study of Llama 3 on three counting tasks.}
\label{tab:case_study_others}
\end{table*}

\begin{figure*}[t!]
     \centering
     %     \begin{subfigure}[b]{\textwidth}
     %     \centering
     %     \includegraphics[width=\linewidth]{icl_strawberry_text.pdf}\caption{Task I: Char Occur.}
     %     \label{fig:icl_strawberry_char_occurrence}
     % \end{subfigure}
    \begin{subfigure}[b]{\textwidth}
         \centering
         \includegraphics[width=\linewidth]{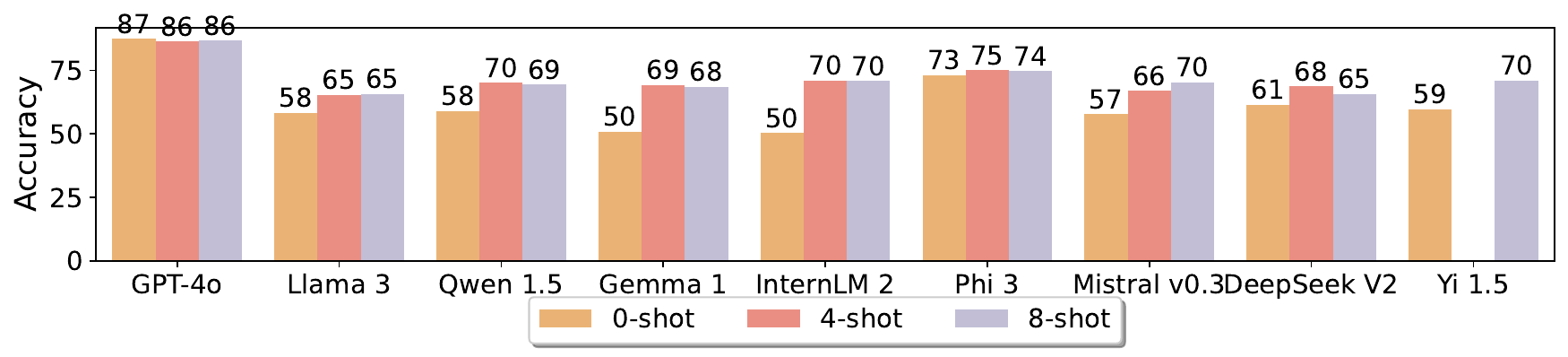}\caption{Task II: Substring Occur.}
         \label{fig:icl_strawberry_substring_occurrence}
     \end{subfigure}
              \begin{subfigure}[b]{\textwidth}
         \centering
         \includegraphics[width=\linewidth]{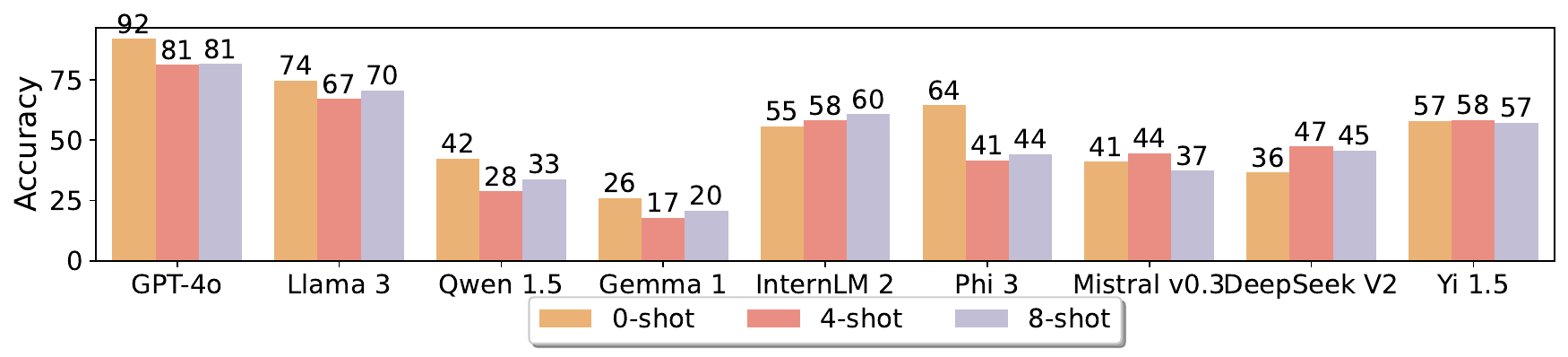}\caption{Task III: Word Len.}
         \label{fig:icl_strawberry_word_length}
     \end{subfigure}
     %     \begin{subfigure}[b]{\textwidth}
     %     \centering
     %     \includegraphics[width=\linewidth]{icl_strawberry_distinct_characters.pdf}\caption{Task IV: Distinct Char.}
     %     \label{fig:icl_strawberry_distinct_characters}
     % \end{subfigure}
        \caption{In-context learning performance of LLMs on Task II and Task III.}
        \label{fig:icl_remains}
        % \vspace{-1em}
\end{figure*}

% \begin{figure*}[t!]
%      \centering
%     % \begin{subfigure}[b]{\textwidth}
%     %      \centering
%     %      \includegraphics[width=\linewidth]{icl_strawberry_text.pdf}\caption{Task I: Char Occur.}
%     %      \label{fig:icl_strawberry_char_occurrence}
%     %  \end{subfigure}
%     % \begin{subfigure}[b]{\textwidth}
%     %      \centering
%          \includegraphics[width=\linewidth]{icl_strawberry_word_length.pdf}%\caption{Task III: Word Len.}
%      %     \label{fig:icl_strawberry_word_length}
%      % \end{subfigure}
%         \caption{In-context learning performance of LLMs on \emph{Task III}. Providing similar examples as demonstrations helps slightly improve performance of open-source models on Task I (~\Cref{fig:icl_strawberry_char_occurrence}) and II (~\Cref{fig:icl_strawberry_substring_occurrence}), while lead performance drop for most models on Task III (this figure) and IV (~\Cref{fig:icl_strawberry_distinct_characters}).}
%         \label{fig:icl}
%         % \vspace{-1em}
% \end{figure*}

\end{document}